%% file: main.tex
\documentclass[final]{siamonline1116}

\input{ex_shared}

\ifpdf
\hypersetup{
  pdftitle={\TheTitle},
  pdfauthor={\TheAuthors}
}
\fi

\begin{document}

\maketitle

\begin{abstract}
  Turbulence-degraded image frames are distorted by both turbulent deformations and space-time-varying blurs. To suppress these effects, we propose a multi-frame reconstruction scheme to recover a latent image from the observed image sequence. Recent approaches are commonly based on registering each frame to a reference image, by which geometric turbulent deformations can be estimated and a sharp image can be restored. A major challenge is that a fine reference image is usually unavailable, as every turbulence-degraded frame is distorted. A high-quality reference image is crucial for the accurate estimation of geometric deformations and fusion of frames. Besides, it is unlikely that all frames from the image sequence are useful, and thus frame selection is necessary and highly beneficial. In this work, we propose a variational model for joint subsampling of frames and extraction of a clear image. A fine image and a suitable choice of subsample are simultaneously obtained by iteratively reducing an energy functional. The energy consists of a fidelity term measuring the discrepancy between the extracted image and the subsampled frames, as well as regularization terms on the extracted image and the subsample. Different choices of fidelity and regularization terms are explored. By carefully selecting suitable frames and extracting the image, the quality of the reconstructed image can be significantly improved. Extensive experiments have been carried out, which demonstrate the efficacy of our proposed model. In addition, the extracted subsamples and images can be put in existing algorithms to produce improved results.
\end{abstract}


\section{Introduction} \label{sec:introduction}
The problem of restoring a clear image from a sequence of turbulence-degraded frames is of high research interest, as the effect of geometric distortions and space-and-time-varying blur would significantly degrade image quality. Under the effects of the turbulent flow of air and changes in temperature, density of air particles, humidity and carbon dioxide level, the refractive index changes accordingly and light is refracted through several turbulence layers \cite{hufnagel1964modulation} \cite{roggemann1996imaging}. Therefore, when we want to capture images in locations where the temperature variation is large, for instance, deserts, roads with tons of vehicles, objects around flames, or from a long distance to perform long-range surveillance or to take pictures of the moon, rays from the objects would arrive at misaligned positions on the imaging plane, and thus distorted images are formed. In general, there are two types of approaches to deal with the problem, one being hardware-based adaptive optics techniques \cite{pearson1976atmospheric} \cite{tyson2015principles} and the other being image-processing-based methods \cite{shimizu2008super} \cite{li2007atmospheric} \cite{vorontsov1999parallel} \cite{seitz2009filter} \cite{hirsch2010efficient}. In this paper, we focus on an image-processing-based method to restore the image. Since we are working on a sequence of distorted images or turbulence-degraded video, we assume the original image is static and the image sensor is also fixed. In order to model this problem, the mathematical model in this paper is based on  \cite{zhu2013removing,furhad2016restoring},\begin{equation} \label{eq1}
I_t(x)=[D_t\big(H_t(I^*)\big)](x)+n_t(x),\indent t=1,\cdots,N
\end{equation}
where $I_t,I^*,$ and $n_t$ are the captured frame at time $t$, the true image, and the sensor noise respectively. The vector $x$ lies in the two-dimensional Euclidean space. $H_t$ represents the blurring operator, which is a space-invariant diffraction-limited point spread function (PSF). $D_t$ is the deformation operator, which is assumed to deform randomly. Note each of the sequences $\{D_t\}$ and $\{n_t\}$ are assumed to be identically distributed random variables, and the subscripts indicate the different actual outcomes that these variables turn out to be at different time instants. Under this formulation, there are three components that need to be tackled, namely the blurring operator $H_t$, the deformation operator $D_t$ and the sensor noise $n_t$.

Most existing restoration methods place their focus on the deformation operator $D_t$. The most intuitive way to reverse $D_t$ is to register each frame to the true image, which, being the solution of the original inverse problem, is unknown beforehand. Hence most approaches to the problem involves estimating the true image with a reference image. However, image registration is computationally costly, and a satisfactory reference image is difficult to obtain from the turbulence-degraded video. This motivates us to look for some efficient methods to compute a clear image quickly, without applying image registration techniques or needing a satisfactory reference image.

Having considered the computational cost of geometric registration-based approaches, while extracting a representative image from turbulence-distorted video, we propose neither to fixate on using deformation-based fidelity nor using the entire sequence. Instead, comparable or even improved results can be achieved by considering other forms of fidelity and a comparatively less-distorted subsampled sequence. We propose adopting a variational model, where various fidelity and regularization terms can be employed to achieve different objectives. As a result, the blurring, deformation and noise effects in the problem setting are simultaneously tackled.

The rest of the paper is organized as follows. In section \ref{sec:previous work}, we review some previous works closely related to this paper. In section \ref{sec:contributions}, we describe the contribution of this paper. In section \ref{sec:proposed algorithm}, our proposed model and algorithm are explained in detail with numerical implementation. We analyze the proposed models and algorithms in section \ref{sec: Analysis}. Experimental results are reported in section \ref{sec:Experimental Result and Discussion}. Finally, we conclude our paper in section \ref{sec:Conclusion}.

\section{Previous work} \label{sec:previous work}

Since the video frames are corrupted by both blur and geometric distortion, it is difficult to deal with them simultaneously, especially in the scenario where a large portion of the images are severely degraded. The registration process is further complicated by the lack of a good reference frame for the observed image sequence. Therefore, several algorithms consisting of registration with reference image and image fusion are proposed.  Meinhardt-Llopis and Micheli \cite{meinhardt2014implementation} \cite{micheli2014linear} proposed a reference extraction method which was coined the centroid method. The basic idea behind is to warp the image by the average deformation field between a fixed image and the others from the turbulence-degraded video. This method assumes the average deformation between the distorted frames and the latent ground truth image to be zero. However, the zero-mean assumption does not hold realistically in the turbulence-distorted video. As a result, it cannot fully resolve the geometric distortion, especially when a large portion of the images are severely degraded. Also, the temporal averaging makes the temporal mean of centroids blurry.

Another method is the ``lucky frame" approach \cite{vorontsov2001anisoplanatic}, which selects the sharpest frame from the video. This method is motivated by statistical proofs \cite{fried1978probability} that given sufficient video frames, there is a high probability that some frames would contain sharp texture details. Since in practice it is rare that one can find a frame which is sharp everywhere, Aubailly {\it et al.} \cite{aubailly2009automated} proposed the Lucky-Region method, which selects at each patch location the sharpest patch across the frames and fuses them together. Anantrasirichai {\it et al.} \cite{anantrasirichai2013atmospheric} adopted this idea and introduced frame selection prior to registration. However, the cost function introduced was coarse, and the selection was done in one step by sorting. As a result, some of the selected frames geometrically differ significantly from the reference image. In addition, the cost function assumed the reference image (i.e. the temporal intensity mean over all frames) to accurately approximate the underlying true image, which is usually not the case. Another similar approach was proposed by Roggemann \cite{roggemann1994image}, where a subsample is selected from images produced by adaptive-optics systems to produce a temporal mean with higher signal-to-noise ratio.

As atmospheric turbulence can severely distort video frames, even if a satisfactory reference image is acquired, the video may not be registered well onto it. A feasible approach to enable registration is to stabilize the video and reduce the deformation between each frame and the reference image. Lou {\it et al.} \cite{lou2013video} proposed to stabilize video by sharpening each frame via spatial Sobolev gradient flow, and temporally smoothing the video to reduce inter-frame deformation. However, the distribution of the image intensities is not preserved under Sobolev gradient sharpening, and temporally smoothing produces ghost artifacts.

Zhu {\it et al.} \cite{zhu2013removing} proposed a B-spline nonrigid registration algorithm to tackle distortion, and a patch-wise temporal kernel regression based near-diffraction-limited (NDL) image restoration to sharpen the image. Finally, they use blind deconvolution algorithm to deblur the fused image. However, the nonrigid registration scheme often misaligns the video frames. Such errors will make the NDL fusion stage further blur the image and produce defects on the fused image. 

Recently, Robust Principal Component Analysis (RPCA) \cite{candes2011robust} is another tool employed to tackle the problem of atmospheric turbulence. He {\it et al.} \cite{he2016atmospheric} proposed a low-rank decomposition approach to separate the registered image sequence into low-rank and sparse parts. The former has less distortion, but is blurry and has few texture details; on the other hand, the latter contains texture information but is noisy. Blind deconvolution is applied on the low-rank part to obtain a deblurred result, which is combined with the enhanced detail layer to get the final result. Xie {\it et al.} \cite{xie2016removing} proposed a hybrid method, which assigns the low-rank image to be the initial reference image. The reference is then improved by solving a variational model, and the frames are registered to the reference image. However, as the deformation between the reference image and the observed frames may be large, direct registration may produce errors.

\section{Contributions} \label{sec:contributions}
The contributions of this work can be summarized as follows:

\medskip

\begin{enumerate}
\item We propose an energy model for joint subsampling of frames and extraction of a restored image from turbulence-degraded video without involving geometric registration. The model produces restored images of comparable or improved quality with other state-of-the-art approaches. 

\item We propose numerical algorithms to iteratively reduce the energies in the models. Experimental results show that the proposed algorithms are effective and highly efficient.

\item We propose different fidelity terms in the energy model. These fidelity terms are carefully explored to investigate their advantages and disadvantages.
\end{enumerate}

\noindent In state-of-the-art methods, costly image registration like optical flow and non-rigid registration are applied, resulting in a very long computational time. Also, since all frames from the video are considered in those algorithms, misalignment occurs in the registration stage for some comparatively severely distorted and blurry frames. As a result, the fusion stage may produce artifacts if the observed video is degraded by severe atmospheric turbulence. In this work, we proposed a variational model to simultaneously obtain an optimal subsample $J$ of frames and extract a reconstructed image $I$. The model is compatible with various fidelity terms and regularization terms. Moreover, effective algorithms are proposed to reduce the energies of the model in order to perform joint subsampling of frames and extraction of a restored image. The proposed method is very flexible that it can tackle severely turbulence-distorted video or even noisy turbulence-degraded video with different regularization terms. The restoration result by the proposed method is dramatically effective that the computational time is within 2 seconds for a turbulence-degraded video with $100$ frames; at the same time they are of comparable quality or even outperform some state-of-the-art methods, which require several thousand seconds or even over ten thousand seconds. Furthermore, the proposed method can serve as a preparatory step for other methods. By applying the proposed extracted image and subsampled video as reference image and video input, the registration process becomes faster (as there are fewer frames in the subsampled video) and more accurate (as a better reference image and video is used), and thus these modified algorithms obtain a more satisfactory result.

\section{Proposed algorithm} \label{sec:proposed algorithm}
In this section, we describe our proposed mathematical model in detail. Our goal is to reconstruct a non-distorted image $I$ from a turbulence-degraded image sequence affected by turbulent deformations and out-of-focus blurs.
\subsection{Proposed model} \label{subsec:proposed model}

Denote a turbulence-degraded image sequence capturing a static object $\mathcal{O}$ by $\mathcal{I} = (I_1,I_2,...,I_n)$. Suppose the size of each image frame $I_k$ is $r\times s$. By stacking each frame $I_k$ as a column vector, $\mathcal{I}$ can be considered as a $rs\times n$ matrix. To restore a sharp and non-distorted image $I$ from $\mathcal{I}$, one commonly used approach is based on a multi-frame reconstruction. This approach is based on registering each frame to a reference image, by which the turbulent deformation matrix can be estimated and a sharp image can be reconstructed. However, one of the main challenges is that a reference image is usually unavailable. A good reference image is necessary for the extraction of turbulent deformations and fusion of image frames. Each frame of the turbulence-distorted video is often degraded by geometric distortions and out-of-focus blurs, which cannot be used as a reference image. On the other hand, frame selection is usually necessary, since it is unlikely that all frames are useful. Therefore, it calls for developing an algorithm which can jointly subsample frames and restore a clear image.

In this work, we propose variational models to simultaneously determine an optimal subsampling $J$ of frames and extract a clear image $I$. Here, $J = \{i_j\in \mathbb{N}: 1\leq i_j\leq n, j=1,2,...,|J|\}$ is the index set representing the subsample of $\mathcal{I}$. Generally speaking, our variational models can be expressed in the following form. We search for $(I,J)$ that minimizes:
\begin{equation} \label{eq:general framework}
E(I,J) = \frac{1}{|J|}\left(\sum_{k\in J} \mathcal{F}(I, I_k) + \lambda \mathcal{Q}(I_k)\right) + \mu \mathcal{R}_1(I) - \tau \mathcal{R}_2 (J) 
\end{equation}

\noindent where $\mathcal{F}$ is the fidelity term measuring the discrepancy between the restored image and frames. $\mathcal{Q}$ is the quality measure of each frame. $\mathcal{R}_1$ and $\mathcal{R}_2$ are the regularization terms for $I$ and $J$ respectively.

There are different choices of $\mathcal{F}$, $\mathcal{Q}$, $\mathcal{R}_1$ and $\mathcal{R}_2$. In this paper, we propose three models for the joint subsampling and restoration of turbulence-degraded images, using different choices of regularization and fidelity terms.

\subsubsection{Model 1} Our goal is to obtain a clear image, which can be treated as a resultant restored image or reference image for the following registration, and a subsampled video which only consists of comparatively sharp and mildly distorted frames. Therefore, Model 1, which is a fast and simple model, is proposed to deal with video mildly and moderately degraded by turbulence.    

In this model, the fidelity term is chosen as the $L^2$-fidelity term, which is commonly used in image restoration. Mathematically, we define $\mathcal{F}(I, I_k) = ||I-I_k||_2^2$. The fidelity term ensures the obtained image to be similar to the images in the subsampled video, which comprises mildly distorted and sharp frames. The quality measure $\mathcal{Q}(I_k)$ of each frame is based on a normalized version of $\Vert \Delta I_k \Vert_1$, i.e.
\begin{equation}
\mathcal Q(I_k)=\dfrac{\max\limits_{i=1,\cdots,n}\Vert\Delta I_i\Vert_1-\Vert\Delta I_k\Vert_1}{\max\limits_{i=1,\cdots,n}\Vert\Delta I_i\Vert_1-\min\limits_{i=1,\cdots,n}\Vert\Delta I_i\Vert_1}
\end{equation}
In essence, $\Delta I$ is the convolution of $I$ with the Laplacian kernel, which captures the features or edges of objects in the image. The magnitude of $\Delta I$ is higher for sharper images.  Hence, $\mathcal Q(I_k)$ is smaller for sharper images. We normalize $\mathcal Q$ to the range of $[0,1]$ for ease of implementation. We have no regularization term $\mathcal{R}_1$ as Model 1 has no additional preference on the restored image. The regularization term $\mathcal{R}_2$ is $1-e^{-\rho |J|}$. This concave increasing function is chosen, because more information can be acquired from more frames, whereas a marginal increase in the size of the subsample has reduced effect on the accuracy of the extracted restored image as the number of subsampled frames increases.

Fixing $J$, the model just obtains the average of the subsampled frames. Therefore, as long as the subsampled frames are sharp and mildly distorted, the resultant restored image is satisfactory. The algorithm details will be illustrated in section \ref{subsec:algorithm 1}.

The overall energy model can be formulated as:
\begin{equation} \label{eq:model1}
E_1(I,J) = \frac{1}{|J|}\sum_{k\in J}\left[||I-I_k||_2^2 + \lambda\mathcal Q(I_k)\right] - \tau(1-e^{-\rho |J|})
\end{equation}
where $\lambda >0$ is a constant for controlling the importance of sharpness of the image frames and $\tau >0$ is a constant for controlling the importance of the number of frames that we want to capture. 
\subsubsection{Model 2} There are some situations where the video is degraded by severe turbulence and all the frames are vigorously distorted and blurry. The simple $L^2$-fidelity term may not accurately measure the similarity between the observed frames and the restored image. Moreover, the restored  image, which is obtained by taking average in the subsampled video, may be locally blurry if the observed video is severely degraded. Therefore, Model 2 is proposed to tackle this situation compromising the computational time but resulting in a more accurate and clear restored image. 

In Model 2, $\mathcal{F}$ is defined as the $L^2$-fidelity between $I$ and the low-rank part of the subsampled frames. More specifically, denote the subsample frames by a $rs\times |J|$ matrix $\mathcal{I}_J$. Robust Principal Component Analysis (RPCA) is applied to $\mathcal{I}_J$. The low-rank part $L$ and the sparse part $S$ are obtained, which captures the general geometric structure and the turbulence respectively. As a result, by fixing $J$ and $L$, the restored image $I$ becomes the average of $L_k$ in the subsampled set $J$. Since the severely turbulence-degraded intensities in the subsampled frames are captured in the sparse component $S$ via RPCA, the restored image $I$ is comparatively clearer and geometrically better-preserved than that obtained in Model 1 with severely turbulence-degraded video. The other terms are the same as Model 1. The algorithm details will be illustrated in section \ref{subsec:algorithm 2}.

The overall energy model can be formulated as
\begin{equation} \label{eq:model2}
E_2(I,J) = \frac{1}{|J|}\left(\sum_{k\in J} ||I-(L_J)_k||_2^2 + \lambda\mathcal Q(I_k)\right) - \tau (1-e^{-\rho |J|}),
\end{equation}
\noindent where $(L_J)_k$ is the $k^\text{th}$ column of $L_J$ and
\begin{equation} \label{eq:RPCA}
(L_J, S_J) = \argmin_{L,S} \{||L||_* + \beta ||S||_1\} \text{ subject to } L+S = \mathcal{I}_J
\end{equation}

\subsubsection{Model 3}  
Our general model is so flexible that the fidelity term and the regularization terms can be changed to suit different needs. To demonstrate the flexibility of the proposed model, a more extreme situation is being tested: noise is added into the real turbulence video. 

In this model, the fidelity term is the $L^2$-fidelity term. The regularization term $\mathcal{R}_1$ is chosen as the total variation (TV) regularization. More specifically, $\mathcal{R}_1 (I) = TV(I)$. By minimizing this term, a clearer and less noisy image can be obtained. The quality measure $\mathcal{Q}(I_k)$ of each frame is also $TV(I_k)$, as less noisy frames are favoured to construct a clear restored image, and it is difficult to estimate sharpness in a noisy image. The regularization term $\mathcal{R}_2$ is $ 1-e^{-\rho |J|}$ as before.

Note that by fixing $J$, this model is similar to the ROF model \cite{rudin1992nonlinear} on a subsampled sequence of images $\{I_k\}_{k\in J}$. The algorithm details will be illustrated in section \ref{subsec:algorithm 3}.

The overall energy model can be formulated as:
\begin{equation} \label{eq:model3}
E_3(I,J) = \frac{1}{|J|}\left(\sum_{k\in J} ||I-I_k||_2^2 + \lambda TV(I_k) \right)+ \mu TV(I) - \tau (1-e^{-\rho |J|}) 
\end{equation}

\subsection{Energy minimization} \label{sec:Energy minimization}
In this subsection, we describe three algorithms to approximate the solutions of the above models, namely \textit{image restoring and image subsampling} (IRIS), \textit{low-rank image restoring and image subsampling} (LIRIS) and \textit{total variation image restoring and image subsampling} (TVIRIS). 

\subsubsection{IRIS algorithm} \label{subsec:algorithm 1}
Given a moderately turbulence-degraded image sequence capturing a static object $\mathcal{O}$ by $\mathcal{I} = (I_1,I_2,...,I_n)$, we now describe a fast and efficient algorithm to obtain a subsampled set $J$ consisting of sharp and mildly distorted frames along with reconstructing a clear restored image $I$ simultaneously, as described by the variational model (\ref{eq:model1}) in the last subsection. Intuitively, this model aims to find the optimal restored image $I$ and subsampled set $J$ simultaneously. $||I-I_k||_2^2$ helps to ensure that the restored image is similar to each $I_k$ in $J$. Each $I_k$ are comparatively sharp among the whole image sequence, which is controlled by $\Vert \Delta I_k \Vert_1$. Traditionally, as much as possible of the observed information should be used to obtain the best result. However, based on the statistical proofs \cite{fried1978probability}, it is not reasonable to assume all the frames in a short exposure with atmospheric turbulence having the same quality. Therefore, we quantify the diminishing improvement with larger samples by the concave increasing function $\tau (1-e^{-\rho |J|})$.

Now, to solve the optimization problem (\ref{eq:model1}), an alternating minimization scheme is applied. Suppose $\lambda$ and $\rho$ are fixed, and an initial image $I^0$ is obtained. Also, the quality measure $\mathcal{Q}(I_k)$ of each frame and the regularization term $\mathcal{R}_2$ for each $|J| \in \{ 1, 2, \ldots, n \}$ are calculated. The initial image $I^0$ is the temporal average of the whole sequence, i.e. 
\begin{equation}\label{eq:temp_mean}
    I^0= \dfrac{1}{n} \sum\limits_{k=1}^n I_k 
\end{equation}
The iterative scheme can then be described as follows for the $t^\text{th}$ iteration:
\begin{enumerate}
    \item Fixing $I^{t-1}$, we minimize $E_1(I^{t-1},J)$ over $\mathcal P(\{1,\cdots,n\})$ to obtain $J^{t}$, i.e. \begin{equation} \label{eq: IRIS I-sub}
        J^t = \argmin_J \frac{1}{|J|}\left(\sum_{k\in J} ||I^{t-1}-I_k||_2^2 + \lambda\mathcal Q(I_k)\right) - \tau (1-e^{-\rho |J|}) 
    \end{equation}
    Note that $||I^{t-1}-I_k||_2^2$ can be calculated easily for each iteration. Also, $\mathcal{R}_2$ and $\mathcal Q(I_k)$ have been calculated before the iteration starts. Denote $||I^{t-1}-I_k||_2^2 + \lambda\mathcal Q(I_k)$ by $E_{1,k}$. Arrange $E_{1,k}$ such that:
    \begin{equation}
        E_{1,k_1}\leq E_{1,k_2}\leq ...\leq E_{1,k_j}\leq ... \leq E_{1,k_n}.
    \end{equation}
    Then denote $S_j$ be the accumulated energy, i.e.
    \begin{equation}
        S_j = \dfrac{1}{j} \left(\sum_{k=1}^j E_{1,k_j} \right) - \tau (1-e^{-\rho j}).
    \end{equation}
    Then arrange $S_j$ such that:
    \begin{equation}
        S_{j_1}\leq S_{j_2}\leq ... \leq S_{j_n}.
    \end{equation}
    Then we obtain the optimal set $J^t$, 
    \begin{equation}
        J^t=\{ k_{1}, k_{2}, \ldots, k_{j_1} \}    
    \end{equation}
    \item Fixing $J^t$, we minimize $E_1(I,J^t)$ over $\mathbb R^{r\times s}$ to obtain $I^t$, i.e.
    \begin{equation} \label{eq: I subproblem model 1 }
        I^t = \argmin_I \frac{1}{|J^t|}\left(\sum_{k\in J^t} ||I-I_k||_2^2 + \lambda\mathcal Q(I_k)\right) - \tau (1-e^{-\rho |J^t|}).
    \end{equation}
    Note that when $J^t$ is fixed, the quality measure $\mathcal{Q}(I_k)$ and the regularization term  $\mathcal{R}_2$ are constant. Hence the $I$-subproblem (\ref{eq: I subproblem model 1 }) becomes 
    \begin{equation}
        I^t = \argmin_I \frac{1}{|J_t|}\left(\sum_{k\in J_t} ||I-I_k||_2^2 \right).
    \end{equation}
    By differentiating with respect to $I$, the minimizer is given by the temporal mean of $\{I_k\}_{k\in J_t}$:
    \begin{equation}
        I^t = \frac{1}{|J_t|}\sum_{k\in J_t} I_k.
    \end{equation}
\end{enumerate}
Repeat step 1 and step 2 above until the difference $ DE_1=E_1^{t-1}-E_1^t$ between the energies at the current and previous steps is smaller than some hyperparameter $\varepsilon$. The overall algorithm is summarized in Algorithm \ref{alg:model1}.

\begin{algorithm}[t]
\begin{algorithmic}[1]
\Require Turbulence-degraded video sequence $\mathcal{I} = (I_1,I_2,...,I_n)$.
\Ensure Subsampled image sequence $\{ I_k\}_{k \in J^\infty}$; Resultant image $I^\infty$.
\State Compute $I^0 = \dfrac{1}{n} \sum\limits_{k=1}^n I_k $;
\State Compute the Quality measure $\mathcal Q(I_k)$ of each frame $\{I_k\}_{k=1}^n$;
\Repeat
    \State Given $I^{t-1}$, $J^{t-1}$. Fix $I^{t-1}$ and obtain $J^{t}$ by solving
    \begin{equation*}
        J^t = \argmin_J \frac{1}{|J|}\left(\sum_{k\in J} ||I^{t-1}-I_k||_2^2 + \lambda(1-\Vert \Delta I_k \Vert_1)\right) - \tau (1-e^{-\rho |J|});
    \end{equation*}
    \State Compute $E_{1,k} = ||I^{t-1}-I_k||_2^2 + \lambda(1-\Vert \Delta I_k \Vert_1)$ for each $k$ and arrange them in ascending order;
    \State Compute accumulated sum $S_j$ for each $j$ and arrange them in ascending order;
    \State $J^t \gets  \{ k_{1}, k_{2}, \ldots, k_{j_1} \} $;
    \State Fix $J^t$ and obtain $I^{t}$ by solving
    \begin{equation*}
        I^t = \argmin_{I} \frac{1}{|J^t|}\sum_{k\in J^t} ||I-I_k||_2^2;
    \end{equation*}
    \State $I^t \gets \frac{1}{|J^t|}\sum\limits_{k\in J^t} I_k$;
\Until{$E_1^{t-1}-E_1^t \leq \varepsilon$};
\State Obtain desirable subsampled image sequence $\{ I_k\}_{k \in J^\infty}$ and resultant image $I^\infty$;
\end{algorithmic}
\caption{Image Restoring and Image Subsampling (IRIS)}
\label{alg:model1}
\end{algorithm}

\subsubsection{LIRIS algorithm} \label{subsec:algorithm 2}
In IRIS algorithm, the simple $L^2$-fidelity term is applied to achieve a fast and satisfactory result. However, the restored image may be locally blurry if the observed images are degraded under severe atmospheric turbulence. In order to achieve a better resultant restored image, the fidelity term in Model 2 is modified to become the $L^2$-fidelity between $I$ and the low-rank part of the subsampled frames. To solve the optimization problem (\ref{eq:model2}), similar to IRIS algorithm, suppose $\lambda$, $\rho$, $\mathcal{Q}(I_k)$, $\mathcal{R}_2(J)$ and the initial image $I^0$ are obtained. The initial image $I^0$ is the temporal average of the low-rank part of the whole sequence, i.e.
\begin{equation}
    I^0=\dfrac{1}{n} \sum_{k=1}^n (L_{\mathcal{I}})_k
\end{equation}
\noindent where $(L_{\mathcal{I}})_k$ is the $k$ column of $L_{\mathcal{I}}$ and $L_{\mathcal{I}}$ is the low-rank part of the whole sequence $\mathcal{I}$, which is obtained in equation (\ref{eq:RPCA}). The iterative scheme can then be described as follows for the $t^\text{th}$ iteration:
\begin{enumerate}
    \item Fixing $I^{t-1}$, we minimize $E_2(I^{t-1},J)$ over $\mathcal P(\{1,\cdots,n\})$ to obtain $J^{t}$, i.e. \begin{equation}
        J^t = \argmin_J \frac{1}{|J|}\left(\sum_{k\in J} ||I^{t-1}-(L_J)_k||_2^2 + \lambda\mathcal Q(I_k)\right) - \tau (1-e^{-\rho |J|}) .
    \end{equation}
    To minimize this subproblem, RPCA should be applied to each of the $2^n$ possible sampling combinations, which is extremely costly. To relax the subproblem, we approximate the above optimization problem with the following:
    \begin{equation}
        J^t = \argmin_J \frac{1}{|J|}\left(\sum_{k\in J} ||I^{t-1}-(L_{\mathcal{I}})_k||_2^2 + \lambda\mathcal Q(I_k)\right) - \tau (1-e^{-\rho |J|}) .
    \end{equation}
    Intuitively, if $\{(L_{\mathcal{I}})_k\}_{k \in J}$ are similar to $I^{t-1}$ in $L^2$ sense, the associated low-rank part $\{(L_J)_k\}_{k \in J}$ are also similar to $I^{t-1}$. The mathematical justification will be shown in section \ref{sec: Analysis}. Therefore, the $J$-subproblem can be done by sorting similar to subsection \ref{subsec:algorithm 1}. 
    \item Fixing $J^t$, we minimize $E_2(I,J_t)$ over $\mathbb R^{r\times s}$ to obtain $I^t$, i.e.
    \begin{equation}
    I^t=\argmin_{I} \dfrac{1}{|J_t|}\sum\limits_{k\in J_t}\Vert I-(L_{J^t})_k\Vert_2^2
    \end{equation}
    Therefore, to obtain $L_{J^t}$, the following optimization scheme is considered:
    \begin{equation} 
        (L_{J^t}, S_{J^t}) = \argmin_{L,S} \{||L||_* + \beta ||S||_1\} \text{ subject to } L+S = \mathcal{I}_{J^t}.
    \end{equation}
    The augmented Lagrangian form of the above optimization problem can be written as follows and solved by Exact Augmented Lagrange Multiplier (EALM) algorithm:
    \begin{equation}
        \mathcal{L}(L,S,\Lambda,\alpha) = \Vert L \Vert_* + \beta \Vert S \Vert_1 + \langle \Lambda , \mathcal I_{J^t}-L-S \rangle + \dfrac{\alpha}{2} \Vert \mathcal I_{J^t}-L-S \Vert_2^2 
    \end{equation}
    where $\Lambda$ is the Lagrange multiplier and $\beta,\alpha$ are the algorithm parameters. For more details, please refer to \cite{lin2010augmented}. The code for EALM is retrieved from \cite{lrslibrary2015}.\\
    
    Then by differentiating with respect to $I$, the minimizer is given by the temporal mean of $\{(L_{J^t})_k\}_{k\in J^t}$:
    \begin{equation}
        I^t = \frac{1}{|J^t|}\sum_{k\in J^t} (L_{J^t})_k.
    \end{equation}
    
\end{enumerate}
Repeat step 1 and step 2 above until the difference $ DE_2=|E_2^{t-1}-E_2^t|$ between the energies at the current and previous steps is smaller than some hyperparameter $\varepsilon$. The overall algorithm is summarized in Algorithm \ref{alg:model2}.

\begin{algorithm}[t]
\begin{algorithmic}[1]
\Require Turbulence-degraded video sequence $\mathcal{I} = (I_1,I_2,...,I_n)$.
\Ensure Subsampled image sequence $\{ I_k\}_{k \in J^\infty}$; Resultant image $I^\infty$.
\State Compute $L_{\mathcal{I}}$ by
\begin{equation*}
    (L_{\mathcal{I}}, S_{\mathcal{I}}) = \argmin_{L,S} \{||L||_* + \beta ||S||_1\} \text{ subject to } L+S = \mathcal{I};
\end{equation*}
\State Compute $$ I^0=\dfrac{1}{n} \sum_{k=1}^n (L_{\mathcal{I}})_k ;$$
\State Compute the Quality measure $\mathcal Q(I_k)$ of each frame $\{I_k\}_{I_k \in \mathcal{I}}$ as in \ref{alg:model1};
\Repeat
    \State Given $I^{t-1}$, $J^{t-1}$. Fix $I^{t-1}$ and obtain $J^{t}$ by solving
    \begin{equation*}
       J^t = \argmin_J \frac{1}{|J|}\left(\sum_{k\in J} ||I^{t-1}-(L_{\mathcal{I}})_k||_2^2 + \lambda(1-\Vert \Delta I_k \Vert_1)\right) - \tau (1-e^{-\rho |J|}) ;
    \end{equation*}
    \State Compute $E_{2,k} = ||I^{t-1}-(L_{\mathcal{I}})_k||_2^2 + \lambda\mathcal Q(I_k)$ for each $k$ and arrange them in ascending order;
    \State Compute accumulated sum $S_j$ for each $j$ and arrange them in ascending order;
    \State $J^t \gets  \{ k_{1}, k_{2}, \ldots, k_{j_1} \} $;
    \State Fix $J^t$ and obtain $L_{J^t}$ by solving
    \begin{equation*} 
        (L_{J^t}, S_{J^t}) = \argmin_{L,S} \{||L||_* + \beta ||S||_1\} \text{ subject to } L+S = \mathcal{I}_{J^t};
    \end{equation*}
    \State Obtain $I^t$ by solving
    \begin{equation*}
        I^t = \argmin_{I} \frac{1}{|J^t|}\sum_{k\in J^t} ||I-(L_{J^t})_k||_2^2;
    \end{equation*}
    \State $I^t \gets \frac{1}{|J^t|}\sum\limits_{k\in J^t} (L_{J^t})_k$;
\Until{$|E_2^{t-1}-E_2^t| \leq \varepsilon$};
\State Obtain desirable subsampled image sequence $\{ I_k\}_{k \in J^\infty}$ and resultant image $I^\infty$;
\end{algorithmic}
\caption{Low-rank Image Restoring and Image Subsampling (LIRIS)}
\label{alg:model2}
\end{algorithm}

\subsubsection{TVIRIS algorithm} \label{subsec:algorithm 3}
Similarly, we can reconstruct satisfactory resultant images and subsampled videos from turbulence-degraded video with additive Gaussian noise by minimizing $E_3(I,J)$ in (\ref{eq:model3}). Taking the temporal average of the whole sequence as the initial image as in (\ref{eq:temp_mean}), the optimization scheme is as follows:
\begin{enumerate}
    \item Fixing $I^{t-1}$, we minimize $E_3(I^{t-1},J)$ over $\mathcal P(\{1,\cdots,n\})$ to obtain $J^t$, i.e.
    \begin{equation}
    J^t=\argmin_J\dfrac{1}{|J|}\left(\sum\limits_{k\in J}\Vert I^{t-1}-I_k\Vert_2^2+\lambda TV(I_k)\right)-\tau(1-e^{-\rho|J|}),
    \end{equation}
    which can be done by sorting similar to subsection \ref{subsec:algorithm 1};
    
    \item Fixing $J^t$, we minimize $E_3(I,J^t)$ over $\mathbb R^{r\times s}$ to obtain $I^t$, i.e.
    \begin{equation}
    I^t=\argmin_{I}\dfrac{1}{|J^t|}\sum\limits_{k\in J^t}\Vert I-I_k\Vert_2^2+\mu TV(I)
    \end{equation}
    
    If we consider the anisotropic total variation
    $$TV_{aniso}(I)=\Vert\nabla_x I\Vert_1+\Vert\nabla_y I\Vert_1,$$
    then the above energy minimization problem can be relaxed by introducing $D_x$ and $D_y$ to split operators, i.e.
    \begin{align}
    (I^t,(D_x)^t,(D_y)^t)
    &=\argmin_{I,D_x,D_y}\dfrac{1}{|J^t|}\sum\limits_{k\in J^t}\Vert I-I_k\Vert_2^2+\mu(\Vert D_x\Vert_1+\Vert D_y\Vert_1)\\
    &\text{subject to }D_x=\nabla_x I\text{ and }D_y=\nabla_y I\nonumber
    \end{align}
    which by the Augmented Lagrangian method can be unconstrained to
    \begin{align}
    (I^t,(D_x)^t,(D_y)^t,(\Lambda_x)^t,(\Lambda_y)^t)=\argmin_{I,D_x,D_y,\Lambda_x,\Lambda_y}
    &\dfrac{1}{|J^t|}\sum\limits_{k\in J^t}\Vert I-I_k\Vert_2^2+\mu(\Vert D_x\Vert_1+\Vert D_y\Vert_1)\\
    &+\langle\Lambda_x,D_x-\nabla_x I\rangle+\langle\Lambda_y,D_y-\nabla_y I\rangle\nonumber\\
    &+\gamma(\Vert D_x-\nabla_x I\Vert_2^2+\Vert D_y-\nabla_y I\Vert_2^2)\nonumber
    \end{align}
    Then the $I$-subproblem is:
    \begin{align}
        I^t=\argmin_{I}
        &\dfrac{1}{|J^t|}\sum\limits_{k\in J^t}\Vert I-I_k\Vert_2^2-\langle(\Lambda_x)^{t-1},\nabla_x I\rangle-\langle(\Lambda_y)^{t-1},\nabla_y I\rangle\\
        &+\gamma(\Vert(D_x)^{t-1}-\nabla_x I\Vert_2^2+\Vert(D_y)^{t-1}-\nabla_y I\Vert_2^2),\nonumber
    \end{align}
    which can be solved with the following linear system
    \begin{equation}
    (Id+\gamma\Delta)I^t=\dfrac{1}{|J^t|}\sum\limits_{k\in J^t}I_k+\dfrac{1}{2}(\nabla_x^*(\Lambda_x)^{t-1}+\nabla_y^*(\Lambda_y)^{t-1})+\gamma(\nabla_x^*(D_x)^{t-1}+\nabla_y^*(D_y)^{t-1}),
    \label{eq:model3anisoI}
    \end{equation}
    where $Id$ is the identity matrix.
    The $D_x$-subproblem is:
    \begin{align}
        (D_x)^t&=\argmin_A\mu\Vert A\Vert_1-\langle(\Lambda_x)^{t-1},A\rangle+\gamma\Vert A-\nabla_x I^t\Vert_2^2\\
        &=\argmin_A\mu\Vert A\Vert_1+\gamma\Vert A-\dfrac{(\Lambda_x)^{t-1}}{2\gamma}-\nabla_x I^t\Vert_2^2\nonumber\\
        &=\argmin_A\sum\limits_{i,j}[\mu|A_{ij}|+\gamma(A_{ij}-\dfrac{\big((\Lambda_x)^{t-1}\big)_{ij}}{2\gamma}-(\nabla_x I^t)_{ij})^2],\nonumber
    \end{align}
    which decouples over space:
    \begin{align}
        \big((D_x)^t\big)_{ij}&=\argmin_x[\mu|x|+\gamma(x-\dfrac{\big((\Lambda_x)^{t-1}\big)_{ij}}{2\gamma}-(\nabla_x I^t)_{ij})^2]\\
        &=\begin{cases}
        \max\{\dfrac{\big((\Lambda_x)^{t-1}\big)_{ij}}{2\gamma}+(\nabla_x I^t)_{ij}-\dfrac{\mu}{2\gamma},0\}&\text{if }(\nabla_x I^t)_{ij}>0\\
        0&\text{if }(\nabla_x I^t)_{ij}=0\\
        \min\{\dfrac{\big((\Lambda_x)^{t-1}\big)_{ij}}{2\gamma}+(\nabla_x I^t)_{ij}+\dfrac{\mu}{2\gamma},0\}&\text{if }(\nabla_x I^t)_{ij}<0
        \end{cases}\nonumber\\
        &=\shrink_{\frac{\mu}{2\gamma}}(\dfrac{\big((\Lambda_x)^{t-1}\big)_{ij}}{2\gamma}+(\nabla_x I^t)_{ij})=\dfrac{1}{2\gamma}\shrink_\mu\Big(\big((\Lambda_x)^{t-1}\big)_{ij}+2\gamma(\nabla_x I^t)_{ij}\Big)\nonumber,
    \end{align}
    and thus
    \begin{equation}
        (D_x)^t=\dfrac{1}{2\gamma}\shrink_\mu((\Lambda_x)^{t-1}+2\gamma(\nabla_x I^t)).
    \label{eq:model3anisoDx}
    \end{equation}
    Similarly, the $D_y$-subproblem yields:
    \begin{equation}
        (D_y)^t=\dfrac{1}{2\gamma}\shrink_\mu((\Lambda_y)^{t-1}+2\gamma(\nabla_y I^t)).
    \label{eq:model3anisoDy}
    \end{equation}
    Finally the multipliers $\Lambda_x$ and $\Lambda_y$ are updated accordingly:
    \begin{align}
        (\Lambda_x)^t&=(\Lambda_x)^{t-1}+\dfrac{1}{2\mu}((D_x)^t-\nabla_x I^t)\\
        (\Lambda_y)^t&=(\Lambda_y)^{t-1}+\dfrac{1}{2\mu}((D_y)^t-\nabla_y I^t)
    \label{eq:model3anisoL}
    \end{align}
    \begin{algorithm}[H]
\begin{algorithmic}[1]
\Require Turbulence-degraded video sequence $\mathcal{I} = (I_1,I_2,...,I_n)$.
\Ensure Subsampled image sequence $\{ I_k\}_{k \in J^\infty}$; Resultant image $I^\infty$.
\State Compute $I^0 = \dfrac{1}{n} \sum\limits_{k=1}^n I_k $;
\State Compute the Quality measure $\mathcal Q(I_k)=TV(I_k)$ of each frame $\{I_k\}_{k=1}^n$;
\Repeat
    \State Given $I^{t-1}$, $J^{t-1}$. Fix $I^{t-1}$ and obtain $J^{t}$ by solving
    \begin{equation*}
        J^t = \argmin_J \frac{1}{|J|}\left(\sum_{k\in J}\Vert I^{t-1}-I_k\Vert_2^2 + \lambda(TV(I_k))\right) - \tau (1-e^{-\rho |J|});
    \end{equation*}
    \State Compute $E_{3,k} = ||I_{t-1}-I_k||_2^2 + \lambda TV(I_k)$ for each $k$ and arrange them in ascending order;
    \State Compute accumulated sum $S_j$ for each $j$ and arrange them in ascending order;
    \State $J^t \gets  \{ k_{1}, k_{2}, \ldots, k_{j_1} \} $;
    \State Fix $J^t$ and obtain $I^{t}$ by solving
    \begin{equation*}
        I^t = \argmin_{I} \frac{1}{|J^t|}\sum_{k\in J^t} ||I-I_k||_2^2 + \mu TV(I);
    \end{equation*}
    \If{$TV=TV_{aniso}$}
    \Repeat
        \State{$I^t\gets\text{Equation }(\ref{eq:model3anisoI})$}
        \State{$(D_x)^t\gets\text{Equation }(\ref{eq:model3anisoDx})$}
        \State{$(D_y)^t\gets\text{Equation }(\ref{eq:model3anisoDy})$}
        \State{$(\Lambda_x)^t,(\Lambda_y)^t\gets\text{Equation }(\ref{eq:model3anisoL})$}
    \Until{$|E_3^{t,m}-E_3^{t,m-1}|\leq\varepsilon$};
    \ElsIf{$TV=TV_{iso}$}
    \Repeat
    \State{$I^t\gets\text{Equation }(\ref{eq:model3isoI})$}
    \State{$(D_x)^t\gets\text{Equation }(\ref{eq:model3isoDx})$}
    \State{$(D_y)^t\gets\text{Equation }(\ref{eq:model3isoDy})$}
    \State{$s_{i,j}^t\gets\sqrt{(\nabla_x I^t)_{ij}^2+(\nabla_y I^t)_{ij}^2}$}
    \State{$(\Lambda_x)^t,(\Lambda_y)^t\gets\text{Equation }(\ref{eq:model3isoL})$}
    \Until{$|E_3^{t,m-1}-E_3^{t,m}|\leq\varepsilon$};
    \EndIf
\Until{$|E_3^{t-1}-E_3^t| \leq \varepsilon$};
\State Obtain desirable subsampled image sequence $\{ I_k\}_{k \in J^\infty}$ and resultant image $I^\infty$;
\end{algorithmic}
\caption{Total Variation Image Restoring and Image Subsampling (TVIRIS)}
\label{alg:model3}
\end{algorithm}
    On the other hand, if we consider the isotropic total variation
    $$TV_{iso}(I)=\sum\limits_{i,j}\sqrt{(\nabla_x I)_{ij}^2+(\nabla_y I)_{ij}^2},$$
    the energy minimization problem can be similarly relaxed by introducing $D_x$ and $D_y$ to split operators, i.e.
    \begin{align}
        (I^t,(D_x)^t,(D_y)^t)
        &=\argmin_{I,D_x,D_y}\dfrac{1}{|J^t|}\sum\limits_{k\in J^t}\Vert I-I_k\Vert^2+\mu\sum\limits_{i,j}\sqrt{(D_x)_{ij}^2+(D_y)_{ij}^2}\\
        &\text{subject to }D_x=\nabla_x I\text{ and }D_y=\nabla_y I \nonumber
    \end{align}
    which by the Augmented Lagrangian Method can be unconstrained to
    \begin{align}
        (I^t,(D_x)^t,(D_y)^t,(\Lambda_x)^t,(\Lambda_y)^t)=\argmin_{I,D_x,D_y,\Lambda_x,\Lambda_y}
        &\dfrac{1}{|J^t|}\sum\limits_{k\in J^t}\Vert I-I_k\Vert^2+\mu\sum\limits_{i,j}\sqrt{(D_x)_{ij}^2+(D_y)_{ij}^2}\\
        &+\langle\Lambda_x,D_x-\nabla_x I\rangle+\langle\Lambda_y,D_y-\nabla_y I\rangle\nonumber\\
        &+\gamma(\Vert D_x-\nabla_x I\Vert_2^2+\Vert D_y-\nabla_y I\Vert_2^2)\nonumber
    \end{align}
    Then the $I$-subproblem can be solved with the same linear system as for the anisotropic case, i.e.
    \begin{equation}
        (Id+\gamma\Delta)I^t=\dfrac{1}{|J^t|}\sum\limits_{k\in J^t}I_k+\dfrac{1}{2}(\nabla_x^*(\Lambda_x)^{t-1}+\nabla_y^*(\Lambda_y)^{t-1})+\gamma(\nabla_x^*(D_x)^{t-1}+\nabla_y^*(D_y)^{t-1}).
    \label{eq:model3isoI}
    \end{equation}
    On the contrary, the $D_x$- and $D_y$-subproblems vary from the anisotropic case, and they no longer decouple over space.\\
    
    The $D_x$-subproblem is:
    \begin{equation}
        (D_x)^t=\argmin_A\mu\sum\limits_{i,j}\sqrt{A_{ij}^2+((D_y)^{t-1})_{ij}^2}+\langle(\Lambda_x)^{t-1},A\rangle+\gamma\Vert A-\nabla_x I^t\Vert_2^2,
    \end{equation}
    whose minimizer $(D_x)^t$ satisfies
    \begin{equation}
        \dfrac{\mu\big((D_x)^t\big)_{ij}}{\sqrt{\big((D_x)^t\big)_{ij}^2+\big((D_y)^{t-1}\big)_{ij}^2}}+\big((\Lambda_x)^{t-1}\big)_{ij}+2\gamma\Big(\big((D_x)^t\big)_{ij}-(\nabla_x I^t)_{ij}\Big)=0,
    \end{equation}
    the first term of which renders the problem nonlinear. Hence we further relax the problem by introducing
    \begin{equation}
        s_{i,j}^t=\sqrt{\big((D_x)^t\big)_{ij}^2+\big((D_y)^t\big)_{ij}^2},
    \end{equation}
    and then explicitly solve the linear equations
    \begin{equation}
        (\mu+2\gamma s_{i,j}^{t-1})((D_x)^t)_{ij}=s_{i,j}^{t-1}\big(2\gamma(\nabla_x I^t\big)_{ij}-\big((\Lambda_x)^{t-1}\big)_{ij}).
    \label{eq:model3isoDx}
    \end{equation}
    Similarly, the $D_y$-subproblem can be solved with the linear equations
    \begin{equation}
        (\mu+2\gamma s_{i,j}^{t-1})\big((D_y)^t\big)_{ij}=s_{i,j}^{t-1}\Big(2\gamma(\nabla_y I^t)_{ij}-\big((\Lambda_y)^{t-1}\big)_{ij}\Big).
    \label{eq:model3isoDy}
    \end{equation}
    Then each $s_{i,j}^t$ is updated with $(D_x)^t$ and $(D_y)^t$.\\
    
    Finally the multipliers $\Lambda_x$ and $\Lambda_y$ are updated accordingly:
    \begin{align}
        (\Lambda_x)^t&=(\Lambda_x)^{t-1}+\dfrac{1}{2\mu}\big((D_x)^t-\nabla_x I^t\big)\\
        (\Lambda_y)^t&=(\Lambda_y)^{t-1}+\dfrac{1}{2\mu}\big((D_y)^t-\nabla_y I^t\big)
    \label{eq:model3isoL}
    \end{align}
    \end{enumerate}

Steps 1 and 2 are repeated until the difference $DE_3=|E_3^{t-1}-E_3^t|$ is smaller than some hyperparameter $\varepsilon$. The overall algorithm is summarized in Algorithm \ref{alg:model3}.

\section{Analysis of the model} \label{sec: Analysis}

\medskip

\begin{theorem}
\label{thm:model1}
Let $\{I^t,J^t\}_{t=1}^{\infty}$ be the sequence obtained by Algorithm \ref{alg:model1}. Then $E_1(I^{t+1},J^{t+1})\\
\leq E_1(I^t, J^t)$ and the scheme stops after finitely many iterations.
\end{theorem}
\begin{proof}
First, $E_1(I,J)$ has a lower bound. When the subsampled set $J$ is fixed, the optimized $I$ is the temporal average over the subsampled set $J$. Since $\{1,2,\cdots,n\}$ is finite and thus its power set is finite, $E_1(I,J)$ has a lower bound over $\mathbb R^{r\times s}\times\mathcal P(\{1,\cdots,n\})$.

Suppose $I^t$ and $J^t$ are obtained. When $I^t$ is fixed, by applying a simple sorting method the global minimizer $J^{t+1}$ is obtained. Therefore, $E_1(I^{t},J^{t+1}) \leq E_1(I^t, J^t)$. When $J^{t+1}$ is fixed, the global minimizer $I^{t+1}$ has an explicit form, which is 
\begin{equation*}
    I^t = \frac{1}{|J^t|}\sum_{k\in J^t} I_k.
\end{equation*}
Therefore, $E_1(I^{t+1},J^{t+1}) \leq E_1(I^{t},J^{t+1}) \leq E_1(I^t, J^t)$. Since $E_1(I,J)$ has a lower bound, $E_1(I^t,J^t)$ is non-increasing over each iteration of Algorithm \ref{alg:model1}. As each $J^t$ is chosen from the finite set $\mathcal P(\{1,2,\cdots,n\})$, Algorithm \ref{alg:model1} stops in finitely many iterations.
\end{proof}
\bigskip

\begin{theorem}
\label{thm:model2a}
Consider the $J$-subproblem in Algorithm \ref{alg:model2} with fixed $|J|=p$.\\
Let $l_p = \max\limits_{i\in J\subseteq\{1,\cdots,n\}:|J|=p}\Vert(L_{\mathcal I})_i - (L_J)_i\Vert_2$, and let $M=\max\limits_{i\in J\subseteq\{1,\cdots,n\}}\Vert I-(L_J)_i\Vert_2$.\\
Let $d_{E,p}$ be the minimum separation distance between energies $E_i=\Vert I-(L_{\mathcal I}^*)_i\Vert_2^2+\mathcal Q(I_i)$.\\
If $l_p<\dfrac{d_{E,p}}{4M}$ at each iteration, then Algorithm \ref{alg:model2} gives the same subsample $J_p^*$ as if the $J^*$-subproblem is solved via exhaustive search over subsamples of cardinality $p$.
\end{theorem}
\begin{proof}
Define the minimum separation distance $d_{E,p}$ between energies $E_i$'s by $d_{E,p}=\min\limits_{\substack{1\leq i,j\leq n\\i\neq j}}|E_i-E_j|$.
By sorting $\{E_i\}$ into $\{E_{i_j}\}$ in ascending order, $d_{E,p}$ is given by the minimum separation distance between consecutive energies, i.e.
\begin{equation}
    d_{E,p}=\min\limits_{1\leq j\leq N-1}(E_{i_{j+1}}-E_{i_j}),
\end{equation}
where $E_{i_j}\leq E_{i_{j+1}}$ for $j=1,2,\cdots,n-1$.\\
Suppose $l_p<\dfrac{d_{E,p}}{4M}$. Given reference image $I$, let $J_p^*$ be the optimal subsample obtained by exhaustive search over subsamples of cardinality $p$. Then for any $i\in J_p^*$,
\begin{align*}
&~~|\Vert I-(L_{\mathcal I})_i\Vert_2^2-\Vert I-(L_{J_p^*})_i\Vert_2^2|\\
&=[\Vert I-(L_{\mathcal I})_i\Vert_2+\Vert I-(L_{J_p^*})_i\Vert_2]\left|\Vert I-(L_{\mathcal I})_i\Vert_2-\Vert I-(L_{J_p^*})_i\Vert_2\right|\\
&\leq 2M\Vert I-(L_{\mathcal I})_i-I+(L_{J_p^*})_i\Vert_2\\
&=2M\Vert(L_{\mathcal I})_i-(L_{J_p^*})_i\Vert_2\leq 2Ml_p<\dfrac{d_{E,p}}{2},
\end{align*}
and thus the sorted order in the $J$-subproblem in Algorithm \ref{alg:model2} with $|J|=p$ is the same as that produced by exhaustive search. Thus with $|J|=p$ fixed, the index set $J_p$ of frames subsampled by Algorithm \ref{alg:model2} is $J_p^*$.
\end{proof}

\begin{theorem}
\label{thm:model2b}
Let $l=\max\limits_{i\in J\subseteq\{1,\cdots,n\}}\Vert(L_\mathcal I)_i-(L_J)_i\Vert_2$, and let $M=\max\limits_{i\in J\subseteq\{1,\cdots,n\}}\Vert I-(L_J)_i\Vert_2$.\\
Let $d_S$ be the minimum separation distance between accumulated energies $S_k$ defined by
\begin{equation}
S_k=\dfrac{1}{k}\sum\limits_{j=1}^k E_{i_j}-\tau(1-e^{-\rho k}),
\end{equation}
where $E_i$ is defined as in Theorem \ref{thm:model2a}, and $E_{i_j}\leq E_{i_{j+1}}$ for $j=1,\cdots,n-1$.\\
If $l<\dfrac{d_S}{4M}$, then Algorithm \ref{alg:model2} gives the same sequence $\{I^t,J^t\}_{t=1}^\infty$ as if the $J$-subproblem is solved via exhaustive search over all subsamples.
\end{theorem}
\begin{proof}
Suppose $l<\dfrac{d_S}{4M}$. Given reference image $I^t$ at the $t^\text{th}$ iteration, let the optimal subsample from exhaustive search over all samples be $J^{t,*}$. Then for each $i_j\in J^{t,*}$, with $E_{i_j}\leq E_{i_{j+1}}$ for $j=1,\cdots,|J^{t,*}|$,
\begin{align*}
    &~~\Big|S_k-\big(\dfrac{1}{k}\sum\limits_{j=1}^k[\Vert I^t-(L_{J^{t,*}})_{i_j}\Vert_2^2+Q(I_{i_j})]-\tau(1-e^{-\rho k})\big)\Big|\\
    &=\dfrac{1}{k}\left|\sum\limits_{j=1}^k(\Vert I^t-(L_{\mathcal I})_{i_j}\Vert_2^2-\Vert I^t-(L_{J^{t,*}})_{i_j}\Vert_2^2)\right|\\
    &\leq\dfrac{1}{k}\sum\limits_{j=1}^k\left(\Vert I^t-(L_{J^{t,*}})_{i_j}\Vert_2+\Vert I^t-(L_{\mathcal I})_{i_j}\Vert_2\right)\left|\Vert I^t-(L_{J^{t,*}})_{i_j}\Vert_2-\Vert I^t-(L_{\mathcal I})_{i_j}\Vert_2\right|\\
    &\leq\dfrac{2M}{k}\sum\limits_{j=1}^k\Vert(L_{J^{t,*}})_{i_j}-(L_{\mathcal I})_{i_j}\Vert_2\leq 2Ml<\dfrac{d_S}{2},
\end{align*}
and thus the sorted order in the $J$-subproblem in Algorithm \ref{alg:model2} is the same as that produced by exhaustive search. Thus the index set $J^t$ of subsampled frames by Algorithm \ref{alg:model2} is $J^{t,*}$. As long as the subsampled frames remain the same, solving the $I$-subproblem in Algorithm \ref{alg:model2} produces the same $I^t$. Hence given $l<\dfrac{d_S}{4M}$, Algorithm \ref{alg:model2} produces the same sequence $\{I^t,J^t\}_{t=1}^\infty$ as if the $J$-subproblem is solved via exhaustive search over all subsamples.
\end{proof}

\begin{theorem}
\label{thm:model3}
Let $\{I^t,J^t\}_{t=1}^{\infty}$ be the sequence obtained by Algorithm \ref{alg:model3}. Then the scheme for the $I$-subproblem converges by the modified Split Bregman algorithm. 
\end{theorem}
\begin{proof}
The $I$-subproblem is written as follows:
\begin{equation*}
    I^{t+1}=\argmin_I\dfrac{1}{|J^t|}\sum\limits_{i\in J^t}\Vert I-I_i\Vert_2^2+\mu TV(I).
\end{equation*}
Note that each of the functionals $\dfrac{1}{|J^t|}\sum\limits_{i\in J^t}\Vert I-I_i\Vert_2^2$ and $\mu TV(I)$ is convex,\\
and that $\dfrac{1}{|J^t|}\sum\limits_{i\in J^t}\Vert I-I_i\Vert_2^2$ is differentiable. Hence from the results of \cite{goldstein2009splitbregman}, the scheme for the $I$-subproblem is of the form of the Split Bregman algorithm and thus converges.
\end{proof}

\begin{figure}[t]
\centering
\subfloat{\includegraphics[height=0.2\textwidth]{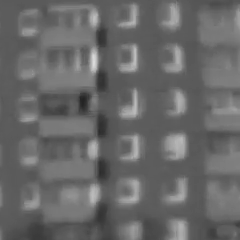}} \hspace{0.1mm}
\subfloat{\includegraphics[height=0.2\textwidth]{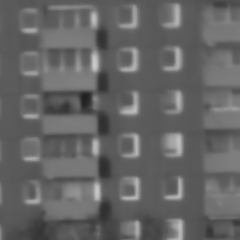}}\hspace{0.1mm} 
\subfloat{\includegraphics[height=0.2\textwidth]{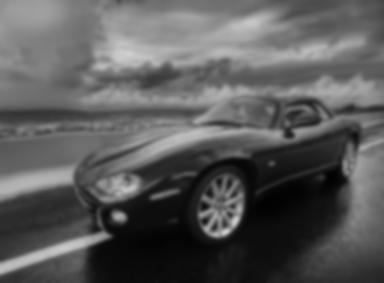}} \hspace{0.1mm}
\subfloat{\includegraphics[height=0.2\textwidth]{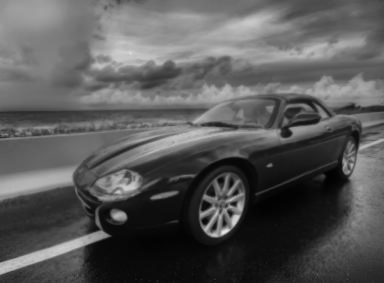}}\hspace{0.1mm} \\ 
\vspace{-3mm}
\subfloat{\includegraphics[height=0.19\textwidth]{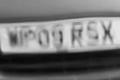}} \hspace{0.1mm}
\subfloat{\includegraphics[height=0.19\textwidth]{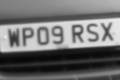}}\hspace{0.1mm}
\subfloat{\includegraphics[height=0.19\textwidth]{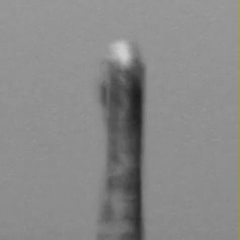}} \hspace{0.1mm}
\subfloat{\includegraphics[height=0.19\textwidth]{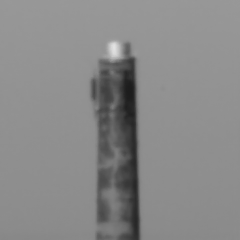}}\hspace{0.1mm}\\ \vspace{-3mm} 
\subfloat{\includegraphics[height=0.18\textwidth]{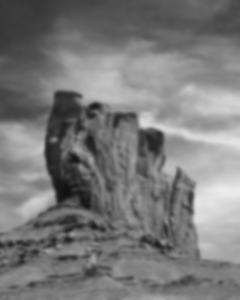}} \hspace{0.1mm}
\subfloat{\includegraphics[height=0.18\textwidth]{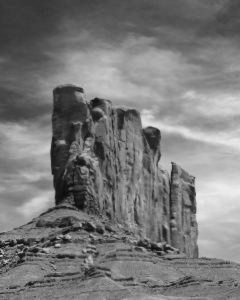}}\hspace{0.1mm}
\subfloat{\includegraphics[height=0.18\textwidth]{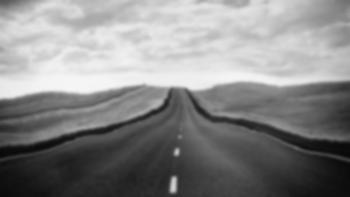}} \hspace{0.1mm}
\subfloat{\includegraphics[height=0.18\textwidth]{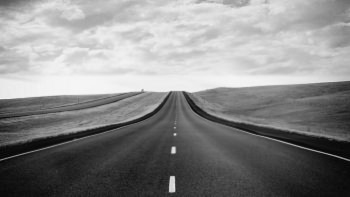}}\hspace{0.1mm} 
\caption{Results of Proposed method. Left: Observed. Right: Proposed method.}
\label{fig:ref compare}
\end{figure}

\section{Experimental Result and Discussion} \label{sec:Experimental Result and Discussion}
In this section, the proposed method is justified in detail and illustratively with experimental results. Firstly, we show the improvement of the final image compared to those of several methods. Both qualitative and quantitative measures are used to evaluate the quality of the restored image by the proposed algorithm compared to several state-of-the-art methods. Peak Signal-to-Noise Ratio (PSNR), Structural Similarity Index (SSIM) and computational time are computed to assess the performance of our proposed method quantitatively. Then, we show the importance of subsampling the video sequence, which not only obtains a better reference image but also reduces the computational time. We also demonstrate situations that motivate the formulation of Models \ref{alg:model2} and \ref{alg:model3}.

To quantitatively evaluate the performance of the proposed algorithm, both simulated data sets (namely Car, Carfront, Desert and Road) and real data sets (namely Building and Chimney) are used to compare with the proposed methods. The Car, Desert and Road sequences are generated with severe simulated turbulence distortions. Each frame of the simulated sequences is generated from a single image by randomly selecting $\frac{\text{width}\times \text{height}}{250} $ positions, and considering an image patch centered at each chosen position. A uniform motion vector patch with the same size of the image patch is generated, whereas the vector is randomly generated from a normal distribution for 2-vectors. Each vector patch is then smoothed with a Gaussian kernel and entrywise multiplied with a distorting strength value. The overall motion vector field is then generated by adding up the vector patches wherever overlapping. The image is then warped by the generated motion vector field. Note that the distortion effects are accumulated where the patches overlap. For each image frame, a Gaussian blur is applied to make them blurry. In the simulated experiments, the chosen patch size is $65 \times 65$, and the mean of the Gaussian kernel is slightly shifted for each image patch. The Desert and Road sequences consist of $100$ frames each, among which 70 frames are degraded under severe distortion and the rest are deformed relatively mildly. The distorting strengths of severely distorted frames are in the range of $[1,1.5]$ while those of mildly distorted frames are in $[0.2,0.3]$. The Carfront sequence is a data set obtained from \cite{anantrasirichai2013atmospheric} which contains mildly distorted frames when compared with the Desert and Road sequences. Note that the Carfront sequence is cropped from the original sequence. The Car sequence contains $80$ frames, among which only 15 are mildly distorted frames and the others are severely distorted. The distorting strengths of the mildly distorted frames and the severely distorted frames are in the ranges of $[0.3,0.5]$ and $[1,1.5]$ respectively. It serves as an extreme test case where most of the frames are severely degraded. As a result, Model \ref{alg:model2} is used for the Car sequence while Model \ref{alg:model1} is used for the other simulated sequences.

For all experiments, the parameters $\lambda$ and $\rho$ in the energy model (\ref{eq:general framework}) in the subsampling stage are in the ranges of $[200, 400]$ and $0.1$ respectively. For Model \ref{alg:model3}, the smoothing parameter is set to be $0.5$. The proposed algorithm is implemented in Matlab with MEX and C++. All the experiments are executed on an Intel Core i7 3.4GHz computer. The error threshold in our experiment is set to be $\varepsilon = 10^{-5}$.

\subsection{Comparison between results of the proposed method with existing methods}
\subsubsection{Quantitative analysis}

\begin{table}[t!]
\centering
\caption{Comparison between the performances of the proposed method and other restoration methods, evaluated with PSNR (in dB), SSIM and computational time (in seconds).}
\label{tab1:psnr ssim}

\begin{tabular}{|l || c | c | c | c | }
\hline  
Sequence                 & SGL & Centroid & NDL & Proposed \\  \hline

\hline
\multirow{3}{*}{Car}   & 23.6366 & 28.3825 & \bf 29.1869 & 28.2689  \\  \cline{2-5}
                        & 0.7558 & 0.8539  & \bf 0.8743 & 0.8624
\\  \cline{2-5}
& \bf 136.8 & 13564 & 10261 & 574.1 \\
\hline
\hline
\multirow{3}{*}{Carfront}   & 16.8052  & 20.1188  & 20.3457 & \bf 20.9223   \\ \cline{2-5}
                        & 0.6920  & 0.8048   & 0.8041  & \bf 0.8375
\\  \cline{2-5}
& 15.5  & 1610.7  & 1793.1  & \bf 1.0535 \\
\hline
\hline
\multirow{3}{*}{Desert}   & 21.1075  & 26.2154  & 23.3818  & \bf 30.2849   \\ \cline{2-5}
                        & 0.7299  & 0.8231   & 0.7605  & \bf 0.9258
\\  \cline{2-5}
& 112.3  & 13778  & 11901  & \bf 1.1621 \\
\hline
\hline
\multirow{3}{*}{Road}   & 24.8007  & 28.1933  & 27.6135  & \bf 32.1232  \\ \cline{2-5}
                        & 0.7479  & 0.8273   & 0.8125 & \bf 0.9005
\\  \cline{2-5}
& 107.3  & 13309  & 11287 & \bf 1.1252 \\
\hline

\end{tabular}
\end{table}

The proposed method is compared with four representative methods: Sobolev gradient-Laplacian method \cite{lou2013video} (\textbf{SGL}), Centroid method \cite{micheli2014linear} (\textbf{Centroid}) and near-diffraction-limited-based image restoration for removing turbulence \cite{zhu2013removing} (\textbf{NDL}). The codes of SGL \cite{lou2013video} and NDL \cite{zhu2013removing} are provided by the respective authors, and the parameters used are default setting. The comparisons are made on results generated from both synthetic sequences, namely Car, Carfront, Desert and Road, and real sequences Building and Chimney.

Table \ref{tab1:psnr ssim} gives the PSNR (in dB), SSIM and the computational time (in seconds) for the restoration results of four different restoration algorithms. The performance indicators of each sequence are contained in three rows, among which the first denotes the PSNR values, the second denotes the SSIM values and the third denotes the computational time. Except for the extreme case, the proposed method demonstrates its effectiveness by outperforming the other methods in PSNR, SSIM and computational time. For the Car sequence which is treated as an extreme case, the proposed method gets a comparable result with Centroid and NDL and outperforms SGL. This is due to the fact that registration is involved in Centroid and NDL.

\subsubsection{Mildly distorted sequences}

\begin{figure}[t!]
\centering
\subfloat[Ground truth]{\includegraphics[width=0.32\textwidth]{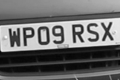}} \hspace{0.01mm}
\subfloat[Observed]{\includegraphics[width=0.32\textwidth]{images/carfront/original.png}}\hspace{0.01mm} 
\subfloat[Proposed]{\includegraphics[width=0.32\textwidth]{images/carfront/I_ref.png}} \\ \vspace{-3mm}
\subfloat[Centroid\cite{micheli2014linear}]{\includegraphics[width=0.32\textwidth]{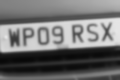}} \hspace{0.01mm}
\subfloat[SGL\cite{lou2013video}]{\includegraphics[width=0.32\textwidth]{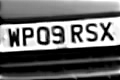}}\hspace{0.01mm} 
\subfloat[NDL\cite{zhu2013removing}]{\includegraphics[width=0.32\textwidth]{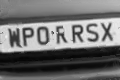}} 
\caption{Comparison between results from the Carfront sequence by the proposed and existing methods}
\label{carfront_proposed}
\end{figure}

The Carfront sequence contains mildly distorted frames only, and the turbulence strengths applied on each image are similar. The restoration results of the Carfront sequence are shown in Figure \ref{carfront_proposed}. The centroid method keeps the geometric structure well but the result is blurry. The shape of the restored image by SGL is slightly distorted and the intensities are unnatural. NDL also keeps the structure relatively well but some artifacts are produced. The proposed method preserves the geometric structure and gets a comparatively sharper result.  

\subsubsection{Strongly distorted sequences}

\begin{figure}[t!]
\centering
\subfloat[Ground truth]{\includegraphics[width=0.32\textwidth]{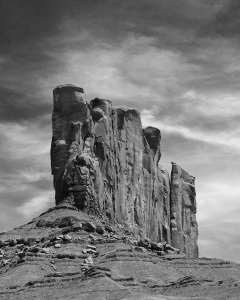}} \hspace{0.01mm}
\subfloat[Observed]{\includegraphics[width=0.32\textwidth]{images/desert/original.png}}\hspace{0.01mm} 
\subfloat[Proposed]{\includegraphics[width=0.32\textwidth]{images/desert/I_ref.png}} \\ \vspace{-3mm}
\subfloat[Centroid\cite{micheli2014linear}]{\includegraphics[width=0.32\textwidth]{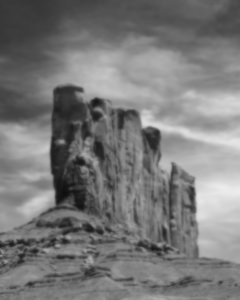}} \hspace{0.01mm}
\subfloat[SGL\cite{lou2013video}]{\includegraphics[width=0.32\textwidth]{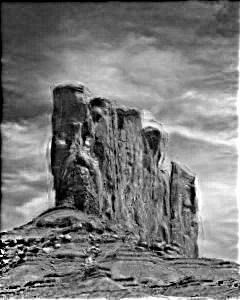}}\hspace{0.01mm} 
\subfloat[NDL\cite{zhu2013removing}]{\includegraphics[width=0.32\textwidth]{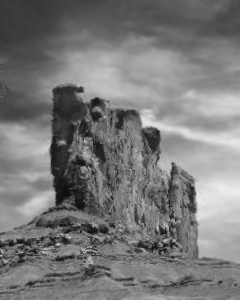}} 
\caption{Comparison between results from the Desert sequence by the proposed and existing methods}
\label{desert_proposed}
\end{figure}

\begin{figure}[t!]
\centering
\subfloat[Ground truth]{\includegraphics[width=0.32\textwidth]{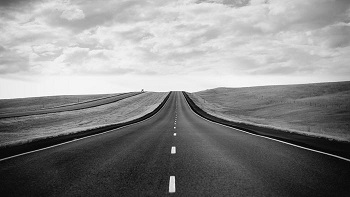}} \hspace{0.01mm}
\subfloat[Observed]{\includegraphics[width=0.32\textwidth]{images/road/original.png}}\hspace{0.01mm} 
\subfloat[Proposed]{\includegraphics[width=0.32\textwidth]{images/road/I_ref.png}} \\ \vspace{-3mm}
\subfloat[Centroid\cite{micheli2014linear}]{\includegraphics[width=0.32\textwidth]{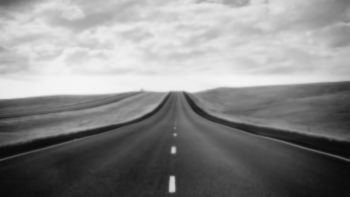}} \hspace{0.01mm}
\subfloat[SGL\cite{lou2013video}]{\includegraphics[width=0.32\textwidth]{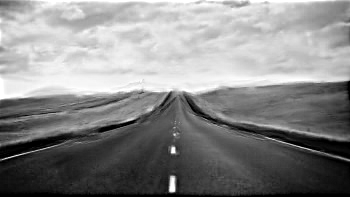}}\hspace{0.01mm} 
\subfloat[NDL\cite{zhu2013removing}]{\includegraphics[width=0.32\textwidth]{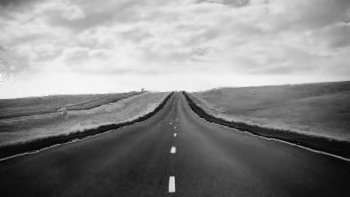}} 
\caption{Comparison between results from the Road sequence by the proposed and existing methods}
\label{road_proposed}
\end{figure}

The majority of the frames in the Desert and Road sequences are strongly distorted, whereas the remaining are mildly distorted. The restoration results of the sequences are shown in Figure \ref{desert_proposed} and Figure \ref{road_proposed}.

Since the deformations among Desert and Road frames are large, the restoration results of the proposed algorithm differ from existing methods. As the imaged objects are significantly displaced across frames, the temporal smoothing effect of the centroid method produces noticeable blur. This is more observable in the Desert experiment, where the many vertical edges are obscured by the blur, whereas in the Road sequence, thin strips parallel to the road are also diminished. A similar temporal smoothing effect manifests in SGL as overlapping shadowy artifacts. Intensity overshoots and jagged edges are observed in the results by NDL, likely because symmetric constraint-based B-spline registration cannot handle random discontinuous displacements across the temporal domain. In comparison, the proposed algorithm preserves clear edges and texture details. It is because the mildly distorted and sharp frames are selected and a good restored image is obtained from these frames. As a result, the proposed algorithm outperforms existing methods. 

\subsubsection{Extreme case}

\begin{figure}[t!]
\centering
\subfloat[Ground truth]{\includegraphics[width=0.32\textwidth]{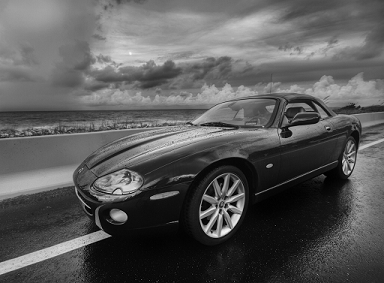}} \hspace{0.01mm}
\subfloat[Observed]{\includegraphics[width=0.32\textwidth]{images/car/original.png}}\hspace{0.01mm} 
\subfloat[Proposed]{\includegraphics[width=0.32\textwidth]{images/car/I_ref.png}} \\ \vspace{-3mm}
\subfloat[Centroid\cite{micheli2014linear}]{\includegraphics[width=0.32\textwidth]{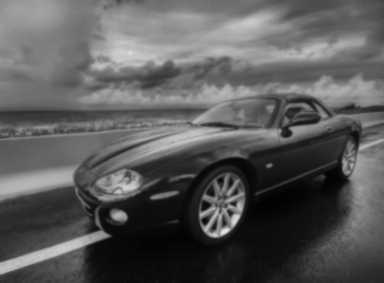}} \hspace{0.01mm}
\subfloat[SGL\cite{lou2013video}]{\includegraphics[width=0.32\textwidth]{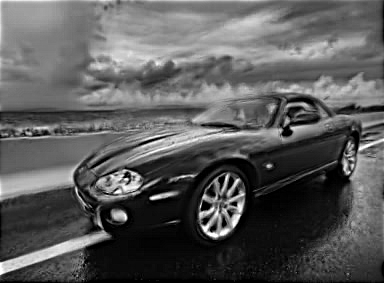}}\hspace{0.01mm} 
\subfloat[NDL\cite{zhu2013removing}]{\includegraphics[width=0.32\textwidth]{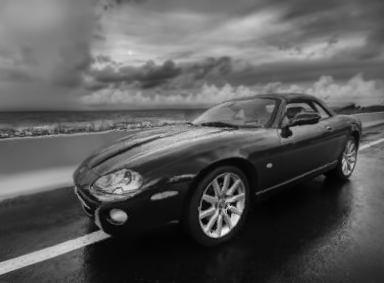}} 
\caption{Comparison between results from the Car sequence by the proposed and existing methods}
\label{car_proposed}
\end{figure}

Most of the frames in the Car sequence are severely distorted, even more so than the Desert and Road sequences. Moreover, the distortions of the mildly distorted frames in the Car sequence are stronger than those in the Desert and Road sequences. The restoration results are shown in Figure \ref{car_proposed}. The result produced by the Centroid method is fairly blurry, and its intensity contrast is significantly lower than other methods. Besides the intensity overshoots, several regions of the SGL result are noticeably deformed. The NDL result has fewer deformed regions. The proposed algorithm preserves the geometric structure and maintains reasonable sharpness. The reason that the proposed restored image is not as sharp as those of the Desert and Road sequences is due to the extremely severe distortion in the Car sequence. Besides, no registration and fusion are involved in the proposed method. Hence the restored image is not as sharp as NDL. 

\subsubsection{Real experiments}
\label{Real experiments}

\begin{figure}[t!]
\centering
\subfloat[Observed]{\includegraphics[width=0.32\textwidth]{images/Building/original.png}}\hspace{0.01mm} 
\subfloat[Proposed]{\includegraphics[width=0.32\textwidth]{images/Building/I_ref.png}} \\ \vspace{-3mm}
\subfloat[Centroid\cite{micheli2014linear}]{\includegraphics[width=0.32\textwidth]{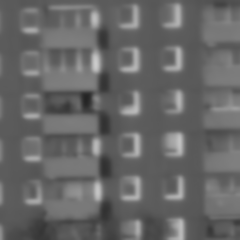}} \hspace{0.01mm}
\subfloat[SGL\cite{lou2013video}]{\includegraphics[width=0.32\textwidth]{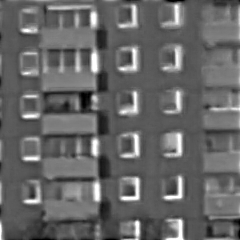}}\hspace{0.01mm} 
\subfloat[NDL\cite{zhu2013removing}]{\includegraphics[width=0.32\textwidth]{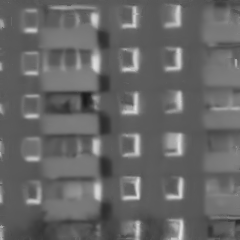}} 
\caption{Comparison between results from the Building sequence by the proposed and existing methods}
\label{Building_proposed}
\end{figure}

\begin{figure}[t!]
\centering
\subfloat[Observed]{\includegraphics[width=0.32\textwidth]{images/chimney/original.png}}\hspace{0.01mm} 
\subfloat[Proposed]{\includegraphics[width=0.32\textwidth]{images/chimney/I_ref.png}} \\ \vspace{-3mm}
\subfloat[Centroid\cite{micheli2014linear}]{\includegraphics[width=0.32\textwidth]{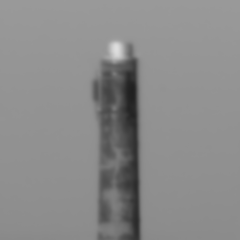}} \hspace{0.01mm}
\subfloat[SGL\cite{lou2013video}]{\includegraphics[width=0.32\textwidth]{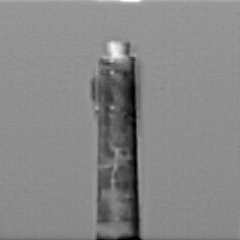}}\hspace{0.01mm} 
\subfloat[NDL\cite{zhu2013removing}]{\includegraphics[width=0.32\textwidth]{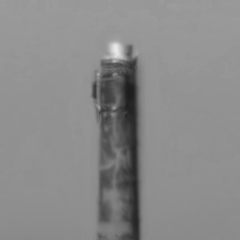}} 
\caption{Comparison between results from the Chimney sequence by the proposed and existing methods}
\label{chimney_proposed}
\end{figure}

We have also tested our proposed method on two real turbulence-distorted sequences, namely the Chimney and Building sequences. The restoration results of the Building sequence and Chimney sequence are shown in Figures \ref{Building_proposed} and \ref{chimney_proposed} respectively. 

The temporal averaging in the centroid method smooths out edges and sharp features as seen in Figure \ref{Building_proposed}(c) and \ref{chimney_proposed}(c). In this aspect, the Sobolev gradient-Laplacian method and near-diffraction-limited method performs better and reconstructs results with sharp details. However, due to varied reasons, the overall intensity distribution of their results differ from that of the original sequence. As a result, the pixel intensities of their results look unnatural. This is exemplified by the presence of dark strips in Figure \ref{Building_proposed}(d), (e) and Figure \ref{chimney_proposed}(d), (e). In addition, some features in the reconstructed results by NDL are visibly distorted in shape. The proposed algorithm preserves the geometric structure better but less sharp than SGL and NDL as there is no fusion stage in the proposed method.

\subsection{Explanation for alternating optimization of subsample and restored image}
\subsubsection{Importance of subsampling} \label{Importance of sub-sampling}

\begin{figure}[t!]
\centering
\subfloat[]{\includegraphics[width=0.24\textwidth]{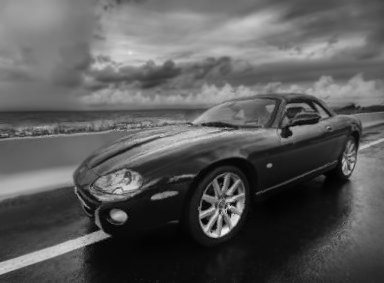}} \hspace{0.01mm}
\subfloat[]{\includegraphics[width=0.24\textwidth]{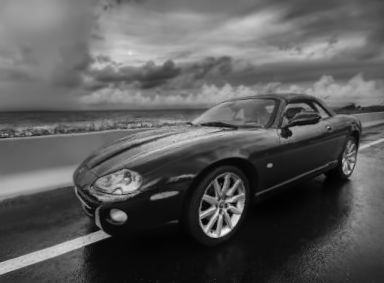}}\hspace{0.01mm} 
\subfloat[]{\includegraphics[width=0.24\textwidth]{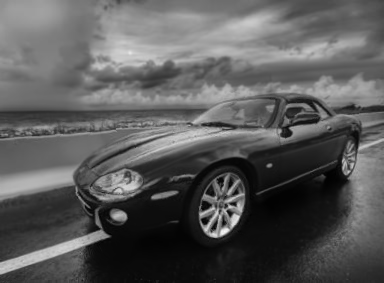}} \hspace{0.01mm}
\subfloat[]{\includegraphics[width=0.24\textwidth]{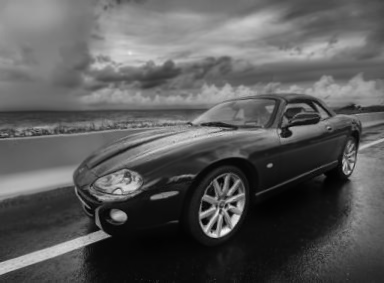}} \\ 
\vspace{-3mm}
\subfloat[]{\includegraphics[width=0.115\textwidth]{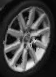}} \hspace{0.01mm}
\subfloat[]{\includegraphics[width=0.115\textwidth]{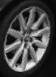}}\hspace{0.01mm} 
\subfloat[]{\includegraphics[width=0.115\textwidth]{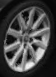}} \hspace{0.01mm}
\subfloat[]{\includegraphics[width=0.115\textwidth]{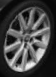}} \hspace{0.01mm}
\subfloat[]{\includegraphics[width=0.115\textwidth]{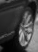}} \hspace{0.01mm}
\subfloat[]{\includegraphics[width=0.115\textwidth]{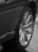}}\hspace{0.01mm} 
\subfloat[]{\includegraphics[width=0.115\textwidth]{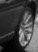}} \hspace{0.01mm}
\subfloat[]{\includegraphics[width=0.115\textwidth]{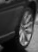}} \\ 
\vspace{-3mm}
\subfloat[]{\includegraphics[width=0.24\textwidth]{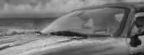}} \hspace{0.01mm}
\subfloat[]{\includegraphics[width=0.24\textwidth]{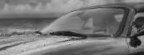}}\hspace{0.01mm} 
\subfloat[]{\includegraphics[width=0.24\textwidth]{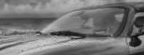}} \hspace{0.01mm}
\subfloat[]{\includegraphics[width=0.24\textwidth]{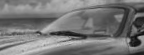}} 
\caption{(a) and (b) are the NDL fusion results from the original Car sequence using temporal mean and the proposed restored image as reference image respectively. (c) and (d) are the NDL fusion result of the subsampled Car sequence using temporal mean and the extracted image we propose as reference image respectively. The PSNR of (a), (b), (c) and (d) are 29.1869, 30.2928, 29.2394 and 30.5191 (in dB) respectively. The computational times including registration and fusion of (a), (b), (c) and (d) are 10261, 10088, 2203 and 2091 (in seconds) respectively. Note that blind deconvolution for deblurring has not been applied to these results.}
\label{fig:importance reference image and subsample}
\end{figure}

In this subsection, the importance of subsampling is demonstrated via qualitative and quantitative measurements. Each frame in a good subsample of the video should have sharp texture details while containing minimal geometric distortion, so that the frames are closely aligned, and as many texture details are kept as possible.

\begin{figure}[t]
\centering
\subfloat{\includegraphics[width=0.19\textwidth]{images/car/original.png}}\hspace{0.01mm} 
\subfloat{\includegraphics[width=0.19\textwidth]{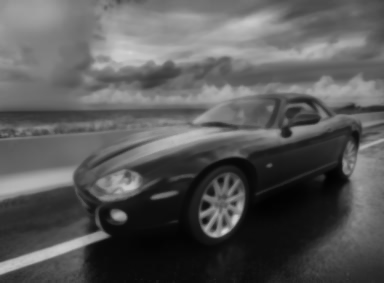}}\hspace{0.01mm}
\subfloat{\includegraphics[width=0.19\textwidth]{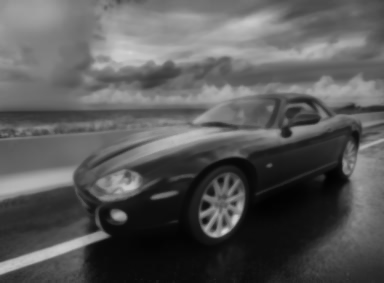}}\hspace{0.01mm}
\subfloat{\includegraphics[width=0.19\textwidth]{images/car/I_ref.png}}
\\ \vspace{-3mm}
\subfloat{\includegraphics[width=0.19\textwidth]{images/carfront/original.png}}\hspace{0.01mm} 
\subfloat{\includegraphics[width=0.19\textwidth]{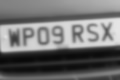}}\hspace{0.01mm}
\subfloat{\includegraphics[width=0.19\textwidth]{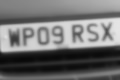}}\hspace{0.01mm}
\subfloat{\includegraphics[width=0.19\textwidth]{images/carfront/I_ref.png}} 
\\ \vspace{-3mm}
\subfloat{\includegraphics[width=0.19\textwidth]{images/desert/original.png}}\hspace{0.01mm} 
\subfloat{\includegraphics[width=0.19\textwidth]{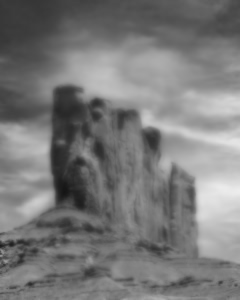}}\hspace{0.01mm}
\subfloat{\includegraphics[width=0.19\textwidth]{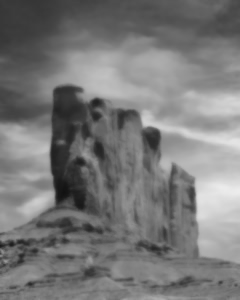}}\hspace{0.01mm}
\subfloat{\includegraphics[width=0.19\textwidth]{images/desert/I_ref.png}}
\\ \vspace{-3mm}
\subfloat{\includegraphics[width=0.19\textwidth]{images/road/original.png}}\hspace{0.01mm} 
\subfloat{\includegraphics[width=0.19\textwidth]{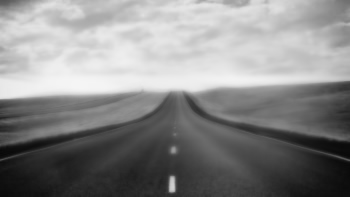}}\hspace{0.01mm} 
\subfloat{\includegraphics[width=0.19\textwidth]{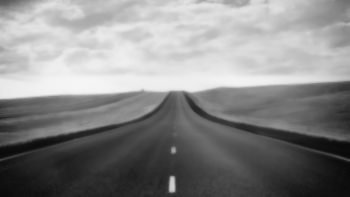}}\hspace{0.01mm}
\subfloat{\includegraphics[width=0.19\textwidth]{images/road/I_ref.png}}
\\ \vspace{-3mm}
\subfloat{\includegraphics[width=0.19\textwidth]{images/Building/original.png}}\hspace{0.01mm}
\subfloat{\includegraphics[width=0.19\textwidth]{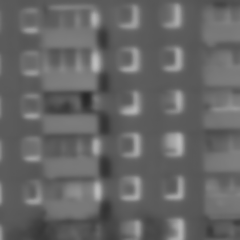}}\hspace{0.01mm} 
\subfloat{\includegraphics[width=0.19\textwidth]{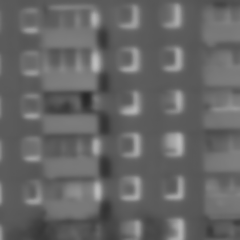}}\hspace{0.01mm}
\subfloat{\includegraphics[width=0.19\textwidth]{images/Building/I_ref.png}}
\\ \vspace{-3mm}
\subfloat{\includegraphics[width=0.19\textwidth]{images/chimney/original.png}}\hspace{0.01mm}
\subfloat{\includegraphics[width=0.19\textwidth]{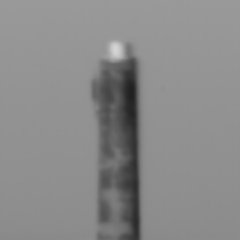}}\hspace{0.01mm}
\subfloat{\includegraphics[width=0.19\textwidth]{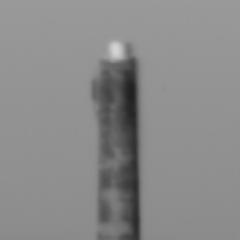}}\hspace{0.01mm}
\subfloat{\includegraphics[width=0.19\textwidth]{images/chimney/I_ref.png}}
\caption{Comparison between the restored images with and without subsampling for the investigated sequences. From left to right are, respectively, an observed frame, the temporal mean used in \cite{zhu2013removing}, the mean of low-rank part by RPCA used in \cite{xie2016removing} and the extracted image of the proposed method.}
\label{fig:reference image}
\end{figure}

\begin{table}[t]
\centering
\caption{Comparison between the restored images with and without subsampling evaluated by PSNR (in dB) and SSIM.}
\label{tab2:reference psnr ssim}

\begin{tabular}{|l || c | c | c |}
\hline  
Sequence & Temporal mean \cite{zhu2013removing} & Mean of low-rank part \cite{xie2016removing} & Proposed\\
\hline
\multirow{2}{*}{Car}   & 25.9726 & 26.6132 & {\bf 28.2689}
\\  \cline{2-4}
                        & 0.7855 & 0.8100 & {\bf 0.8624}
\\
\hline
\hline
\multirow{2}{*}{Carfront}   & 19.3090 & 19.5607 & {\bf 20.9233}
\\ \cline{2-4}
                        & 0.7666 & 0.7770 & {\bf 0.8375}
\\
\hline
\hline
\multirow{2}{*}{Desert}   & 23.9504 & 24.6356 & {\bf 30.2849}
\\ \cline{2-4}
                        & 0.7219 & 0.7506 & {\bf 0.9258}
\\
\hline
\hline
\multirow{2}{*}{Road}   & 25.9164 & 26.9449 & {\bf 32.1232}
\\ \cline{2-4}
                        & 0.7600 & 0.7881 & {\bf 0.9005}
\\
\hline

\end{tabular}
\end{table}

Moreover, note the short computational time of our algorithm (in Table \ref{tab1:psnr ssim}), and the shorter length of the subsampled video compared to original footage. If the subsampled sequence is applied in existing restoration or stabilization algorithms,  the total computational time is reduced significantly. We incorporate Zhu's NDL algorithm \cite{zhu2013removing} to support our claims. The fusion results with and without subsampling are compared. Each video sequence is registered to their corresponding reference image, which is the temporal mean of the sequence. Then fusion is applied to the two registered video sequences. Computational time, visual comparison and quantitative measures will be used to justify our conclusion. Comparing (a) to (c) in Figure \ref{fig:importance reference image and subsample}, noticeable artifacts can be observed in the center of the wheel and the overall image is also blurry (See the zoomed parts in Figure \ref{fig:importance reference image and subsample}(e) and (g)). In contrast to the fusion results of the original video, the wheels in the NDL results obtained by fusing the subsampled sequence are free of artifacts, are sharper and have clearer edges. This observation can be explained by two factors: 
\begin{enumerate}
    \item Since the subsampled video is obtained by maximizing an energy that depends on the number of frames in the subsampled sequence, their similarity to the reference image, and their sharpness, the subsampled image frames mainly consist of comparatively sharp and less deformed image frames. Fewer noisy components are included in the sparse part in the fusion stage, and hence the result has sharper edges and richer texture details are preserved.
    \item Since the reference image is constructed by a sharper and mildly distorted video sequence, the reference image is sharper and better preserves geometric structure. Therefore the alignments of the registered frames are more accurate, and the frames are thus more similar to the reference image. The fusion artifacts due to poor registration become insignificant. 
\end{enumerate} 
Moreover, if both the reference image and the subsampled sequence are taken into account, the best result is achieved. Note that the total time taken to obtain the proposed reference and subsampled sequence, and then using the NDL method on the subsampled sequence with the proposed reference as a reference image, is only 2091 seconds, which is about a fifth of that of the original method (10261 seconds). Adopting the reference image and subsampled sequence of the proposed method yields the best result in computational time, visual quality and quantitative measures, even when using an existing registration and fusion scheme.

\subsubsection{Comparison between the proposed restored image and the reference images employed in other methods} \label{Importance of reference image}

\begin{figure}[t!]
\centering
\subfloat[]{\includegraphics[width=0.19\textwidth]{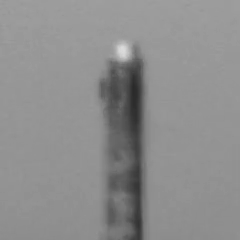}} \hspace{0.01mm}
\subfloat[]{\includegraphics[width=0.19\textwidth]{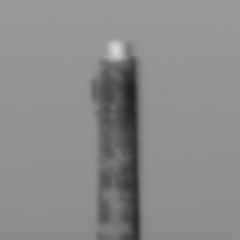}}\hspace{0.01mm} 
\subfloat[]{\includegraphics[width=0.19\textwidth]{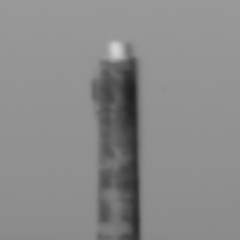}} \hspace{0.01mm}
\subfloat[]{\includegraphics[width=0.19\textwidth]{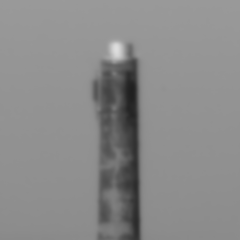}}\hspace{0.01mm} 
\subfloat[]{\includegraphics[width=0.19\textwidth]{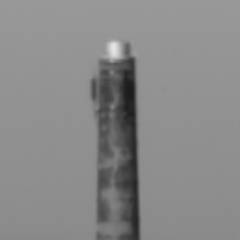}} \\ 
\vspace{-3mm}
\subfloat[]{\includegraphics[width=0.19\textwidth]{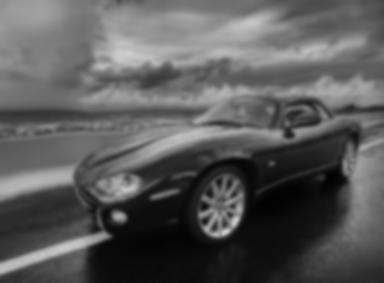}} \hspace{0.1mm}
\subfloat[]{\includegraphics[width=0.19\textwidth]{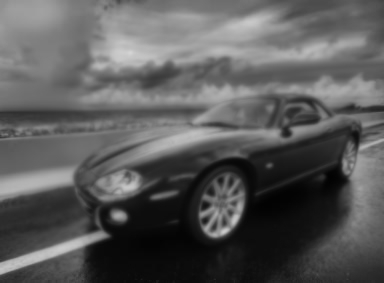}}\hspace{0.1mm} 
\subfloat[]{\includegraphics[width=0.19\textwidth]{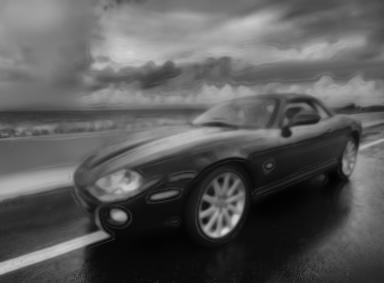}} \hspace{0.1mm}
\subfloat[]{\includegraphics[width=0.19\textwidth]{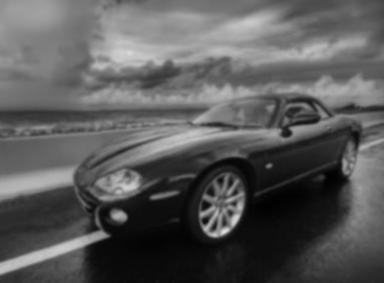}}\hspace{0.1mm} 
\subfloat[]{\includegraphics[width=0.19\textwidth]{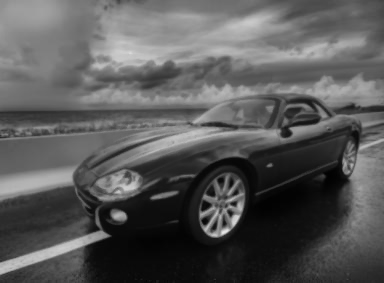}}
\caption{The images extracted from the Chimney and Car sequences. (a)(f) Observed. (b)(g) Temporal mean \cite{zhu2013removing}. (c)(h) Mean of low-rank by RPCA \cite{xie2016removing}. (d)(i) Centroid method \cite{micheli2014linear}. (e)(j) Proposed method. The PSNR of (g), (h), (i), (j) from Car sequence are  
25.9726, 26.3510, 26.7398, 28.2659 (in dB) respectively. Note that blind deconvolution for deblurring has not been applied to these results.}
\label{fig:reference image2}
\end{figure}

Our proposed method is efficient and gives a good reference image. To justify this, both qualitative and quantitative assessments are employed. The qualitative justification of the proposed method is shown by comparing the extracted reference images obtained by the proposed method with those used by other methods. See Figure \ref{fig:reference image}. The effectiveness of the proposed method is justified quantitatively in Table \ref{tab2:reference psnr ssim}. The visual quality of the reference images obtained by the proposed algorithm, temporal averaging, the temporal average of the low-rank and the centroid method \cite{micheli2014linear} are compared qualitatively in this subsection. The reference images are shown in Figure \ref{fig:reference image2}: the first column contain observed images from `Chimney' and `Car' sequences while the other four columns are the reference images generated by temporal mean, mean of low rank, the centroid method and the proposed algorithm. In the Chimney sequence, the proposed algorithm preserves sharpness and details better than the other three methods. This is because the subsampled sequence only consists of sharp and mildly distorted images, and hence the obtained image is clearer. For the other methods, the blurry and severely deformed frames are also taken into account, so the reference image is corrupted. For an even more severely turbulence-degraded video (Car sequence), the blurring effect is more noticeable. From the mean of the low-rank part, the general geometric structure is extracted and so sharp edges are preserved. However, most texture details will go to the sparse part, so the details are removed. For the centroid method, the texture details are kept as every image is warped by a deformation field towards the `average position', and there is no direct manipulation on image intensities except for interpolation. However, since the centroid method is based on the strong zero-mean assumption of the deformation fields between ground truth and the distorted sequence, which does not usually hold for turbulence-distorted video, the geometric structure may not be well kept. For the proposed method, the reference image is reconstructed from a good subsampled sequence, which minimizes the energy (\ref{eq:general framework}) considering similarity and sharpness and is improved iteratively. As a result, the edges are sharp, the geometric structure is preserved and the texture details are kept. The PSNR of the reference images also justifies the result. 

Moreover, the importance of the reference image is tested by comparing the results of the NDL fusion algorithm using a temporal average reference image and our proposed reference image. Original video without subsampling is applied. From Figure \ref{fig:importance reference image and subsample}, it is observable that there are artifacts in the wheel (e, i) and the background (m) of the fused image car with blurry temporal averaging reference image while there are no noticeable artifacts in that with the proposed reference image. The reason behind is that the registration of the image sequence will be more accurate if the reference image is clearer. As a result, the fusion result with the reference image by our proposed method is better in both visual quality and quantitative measure compared with the original NDL method. See Figure \ref{fig:importance reference image and subsample}(a) and (b) and their corresponding zoomed parts.

\subsubsection{How the proposed algorithm can enhance existing methods}

\begin{figure}[t]
\centering
\subfloat[Ground truth]{\includegraphics[width=0.24\textwidth]{images/carfront/groundtruth.png}} \hspace{0.01mm}
\subfloat[Observed]{\includegraphics[width=0.24\textwidth]{images/carfront/original.png}}\hspace{0.01mm} 
\subfloat[Centroid\cite{micheli2014linear}]{\includegraphics[width=0.24\textwidth]{images/carfront/C_o.png}} \hspace{0.01mm}
\subfloat[Centroid\cite{micheli2014linear} + Proposed]{\includegraphics[width=0.24\textwidth]{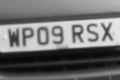}} \\ \vspace{-3mm}
\subfloat[SGL\cite{lou2013video}]{\includegraphics[width=0.24\textwidth]{images/carfront/lou_original.png}} \hspace{0.01mm}
\subfloat[SGL\cite{lou2013video} + Proposed]{\includegraphics[width=0.24\textwidth]{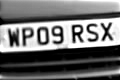}}\hspace{0.01mm} 
\subfloat[NDL\cite{zhu2013removing}]{\includegraphics[width=0.24\textwidth]{images/carfront/img_ndl_o.png}} \hspace{0.01mm}
\subfloat[NDL\cite{zhu2013removing} + Proposed]{\includegraphics[width=0.24\textwidth]{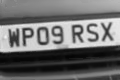}} 
\caption{Comparison between results from the Carfront sequence}
\label{carfront}
\end{figure}

\begin{figure}[t]
\centering
\subfloat[Ground truth]{\includegraphics[width=0.24\textwidth]{images/desert/groundtruth.png}} \hspace{0.01mm}
\subfloat[Observed]{\includegraphics[width=0.24\textwidth]{images/desert/original.png}}\hspace{0.01mm} 
\subfloat[Centroid\cite{micheli2014linear}]{\includegraphics[width=0.24\textwidth]{images/desert/C_o.png}} \hspace{0.01mm}
\subfloat[Centroid\cite{micheli2014linear} + Proposed]{\includegraphics[width=0.24\textwidth]{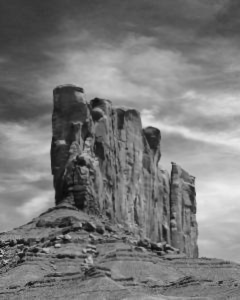}} \\ \vspace{-3mm}
\subfloat[SGL\cite{lou2013video}]{\includegraphics[width=0.24\textwidth]{images/desert/lou_original.png}} \hspace{0.01mm}
\subfloat[SGL\cite{lou2013video} + Proposed]{\includegraphics[width=0.24\textwidth]{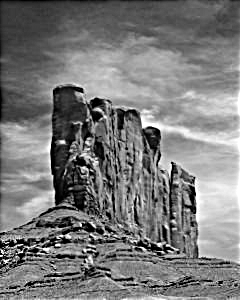}}\hspace{0.01mm} 
\subfloat[NDL\cite{zhu2013removing}]{\includegraphics[width=0.24\textwidth]{images/desert/img_ndl_o.png}} \hspace{0.01mm}
\subfloat[NDL\cite{zhu2013removing} + Proposed]{\includegraphics[width=0.24\textwidth]{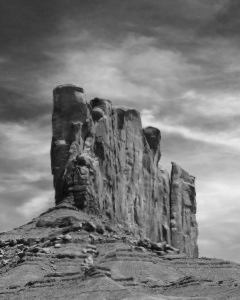}} 
\caption{Comparison between results from the Desert sequence}
\label{desert}
\end{figure}

\begin{figure}[t]
\centering
\subfloat[Ground truth]{\includegraphics[width=0.24\textwidth]{images/road/groundtruth.png}} \hspace{0.01mm}
\subfloat[Observed]{\includegraphics[width=0.24\textwidth]{images/road/original.png}}\hspace{0.01mm} 
\subfloat[Centroid\cite{micheli2014linear}]{\includegraphics[width=0.24\textwidth]{images/road/C_o.png}} \hspace{0.01mm}
\subfloat[Centroid\cite{micheli2014linear} + Proposed]{\includegraphics[width=0.24\textwidth]{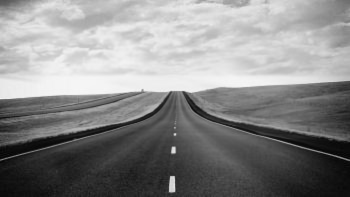}} \\ \vspace{-3mm}
\subfloat[SGL\cite{lou2013video}]{\includegraphics[width=0.24\textwidth]{images/road/lou_original.png}} \hspace{0.01mm}
\subfloat[SGL\cite{lou2013video} + Proposed]{\includegraphics[width=0.24\textwidth]{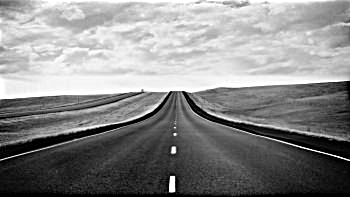}}\hspace{0.01mm} 
\subfloat[NDL\cite{zhu2013removing}]{\includegraphics[width=0.24\textwidth]{images/road/img_ndl_o.png}} \hspace{0.01mm}
\subfloat[NDL\cite{zhu2013removing} + Proposed]{\includegraphics[width=0.24\textwidth]{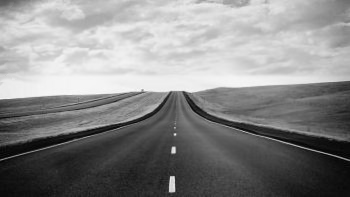}} 
\caption{Comparison between results from the Road sequence}
\label{road}
\end{figure}

\begin{table}[t]
\centering
\caption{Comparison between the performances existing restoration methods of with and without enhancing with the proposed method, evaluated by PSNR (in dB), SSIM and computational time (in seconds).}
\label{tab2:psnr ssim}

\begin{tabular}{|l || c | c | c | c | c | c |  }
\hline  
\multirow{2}{*}{Sequence}                 & \multirow{2}{*}{SGL} & SGL  & \multirow{2}{*}{Centroid} & Centroid  & \multirow{2}{*}{NDL} & NDL   \\ 
&  & + Proposed &  &  + Proposed &  &  + Proposed  \\  \hline

\hline
\multirow{3}{*}{Car}   & 23.6366 & 22.9945 & 28.3825 & 29.7958 & 29.1869 &  30.5191   \\  \cline{2-7}
                        & 0.7558 & 0.7609 & 0.8539 & 0.8867  & 0.8743 &  0.8960 
\\  \cline{2-7}
&  136.8 & 606.2 & 13564 & 1232.3 & 10261 & 2665.2  \\
\hline
\hline
\multirow{3}{*}{Carfront}   & 16.8052 & 17.7973 & 20.1188 & 20.9859 & 20.3457 &  22.2137    \\ \cline{2-7}
                        & 0.6920 & 0.7214 & 0.8048 & 0.8419  & 0.8041 &  0.8665 
\\  \cline{2-7}
& 15.5 & 4.2 & 1610.7 & 65.1 & 1793.1 & 346.0  \\
\hline
\hline
\multirow{3}{*}{Desert}   & 21.1075 & 21.9822 & 26.2154 & 30.3749 & 23.3818 &  30.4087    \\ \cline{2-7}
                        & 0.7299 & 0.7941 & 0.8231 & 0.9273  & 0.7605 &  0.9292 
\\  \cline{2-7}
& 112.3 & 10.5 & 13778 & 78.8 & 11901 & 917.1  \\
\hline
\hline
\multirow{3}{*}{Road}   & 24.8007 & 24.8087 & 28.1933 & 32.3209 & 27.6135 &  33.0020  \\ \cline{2-7}
                        & 0.7479 & 0.7974 & 0.8273 & 0.9013  & 0.8125 &  0.9040 
\\  \cline{2-7}
& 107.3 & 14.2 & 13309 & 177.6 & 11287 & 1344.7  \\
\hline

\end{tabular}
\end{table}

\begin{figure}[t]
\centering
\subfloat[Ground truth]{\includegraphics[width=0.24\textwidth]{images/car/groundtruth.png}} \hspace{0.01mm}
\subfloat[Observed]{\includegraphics[width=0.24\textwidth]{images/car/original.png}}\hspace{0.01mm} 
\subfloat[Centroid\cite{micheli2014linear}]{\includegraphics[width=0.24\textwidth]{images/car/C_o.png}} \hspace{0.01mm}
\subfloat[Centroid\cite{micheli2014linear} + Proposed]{\includegraphics[width=0.24\textwidth]{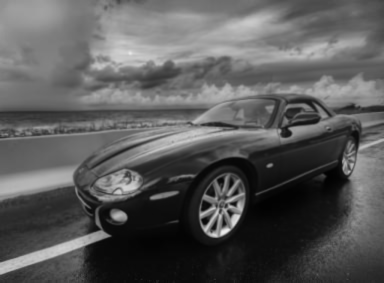}} \\ \vspace{-3mm}
\subfloat[SGL\cite{lou2013video}]{\includegraphics[width=0.24\textwidth]{images/car/lou_original.png}} \hspace{0.01mm}
\subfloat[SGL\cite{lou2013video} + Proposed]{\includegraphics[width=0.24\textwidth]{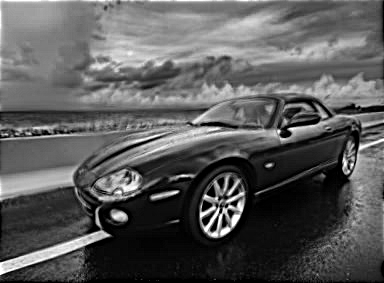}}\hspace{0.01mm} 
\subfloat[NDL\cite{zhu2013removing}]{\includegraphics[width=0.24\textwidth]{images/car/img_ndl_o.png}} \hspace{0.01mm}
\subfloat[NDL\cite{zhu2013removing} + Proposed]{\includegraphics[width=0.24\textwidth]{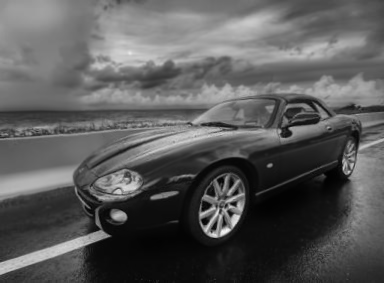}} 
\caption{Comparison between results from the Car sequence}
\label{car}
\end{figure}

\begin{figure}[t]
\centering
\subfloat[Observed]{\includegraphics[width=0.24\textwidth]{images/Building/original.png}}\hspace{0.01mm} 
\subfloat[Centroid\cite{micheli2014linear}]{\includegraphics[width=0.24\textwidth]{images/Building/C_o.png}} \hspace{0.01mm}
\subfloat[Centroid\cite{micheli2014linear} + Proposed]{\includegraphics[width=0.24\textwidth]{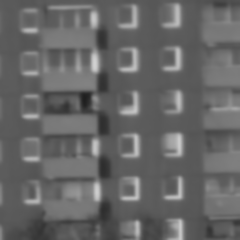}} \\ \vspace{-3mm}
\subfloat[SGL\cite{lou2013video}]{\includegraphics[width=0.24\textwidth]{images/Building/lou_original.png}} \hspace{0.01mm}
\subfloat[SGL\cite{lou2013video} + Proposed]{\includegraphics[width=0.24\textwidth]{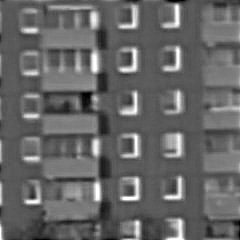}}\hspace{0.01mm} 
\subfloat[NDL\cite{zhu2013removing}]{\includegraphics[width=0.24\textwidth]{images/Building/img_ndl_o.png}} \hspace{0.01mm}
\subfloat[NDL\cite{zhu2013removing} + Proposed]{\includegraphics[width=0.24\textwidth]{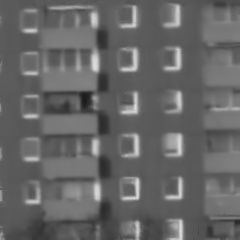}} 
\caption{Comparison between results from the Building sequence}
\label{Building}
\end{figure}

\begin{figure}[t]
\centering
\subfloat[Observed]{\includegraphics[width=0.24\textwidth]{images/chimney/original.png}}\hspace{0.01mm} 
\subfloat[Centroid\cite{micheli2014linear}]{\includegraphics[width=0.24\textwidth]{images/chimney/C_o.png}} \hspace{0.01mm}
\subfloat[Centroid\cite{micheli2014linear} + Proposed]{\includegraphics[width=0.24\textwidth]{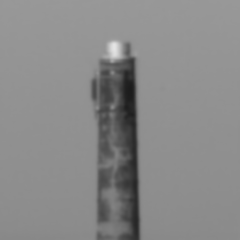}} \\ \vspace{-3mm}
\subfloat[SGL\cite{lou2013video}]{\includegraphics[width=0.24\textwidth]{images/chimney/lou_original.png}} \hspace{0.01mm}
\subfloat[SGL\cite{lou2013video} + Proposed]{\includegraphics[width=0.24\textwidth]{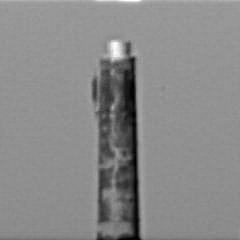}}\hspace{0.01mm} 
\subfloat[NDL\cite{zhu2013removing}]{\includegraphics[width=0.24\textwidth]{images/chimney/img_ndl_o.png}} \hspace{0.01mm}
\subfloat[NDL\cite{zhu2013removing} + Proposed]{\includegraphics[width=0.24\textwidth]{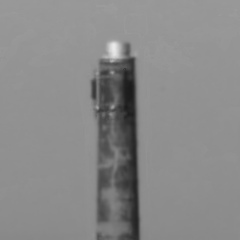}} 
\caption{Comparison between results from the Chimney sequence}
\label{chimney}
\end{figure}

In order to get the best restored image from the turbulence-degraded sequence, registration and fusion are needed. However, in general, there are inevitable drawbacks in the registration process:
\begin{enumerate}
	\item Registration is typically computationally heavy, especially in the context of registering severely distorted sequences with a large number of frames.
	\item A sharp reference image with details and geometric structure preserved is needed in the registration process. Otherwise, misalignment artifacts will be produced in the fusion stage.
\end{enumerate}
The proposed method is not only treated as a restoration method, but also an enhancement to the existing restoration approaches. First, mildly distorted frames are subsampled in the proposed method. This greatly reduces the computational time for registration. Second, a good restored image is obtained by the proposed method so registration is improved. Experiments are carried out and the results are evaluated both qualitatively and quantitatively. From the Table \ref{tab2:psnr ssim}, comparing the results of the proposed method to existing methods, the performance is significantly improved in terms of PSNR, SSIM and computational time. Note that it is reasonable the NDL + Proposed method gets the best result in four sequences, as NDL uses both the subsampled video and restored image. On the other hand, except the SGL case in the Car sequence where the proposed method with Model 3 is applied, all the restoration results by the state-of-the-art methods are improved by applying the proposed method in terms of PSNR, SSIM and computational time. This justifies that the proposed method is dramatically effective. 

\subsection{Justification for the variants of the proposed model}
\begin{figure}[t]
\centering
\subfloat[]{\includegraphics[width=0.32\textwidth]{images/car_original_10.png}} \hspace{0.01mm}
\subfloat[]{\includegraphics[width=0.32\textwidth]{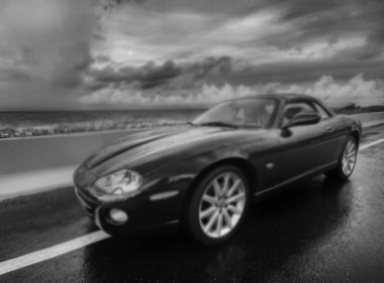}}\hspace{0.01mm} 
\subfloat[]{\includegraphics[width=0.32\textwidth]{images/car_ref_rpca.png}} \\ 
\vspace{-3mm}
\subfloat[]{\includegraphics[width=0.32\textwidth]{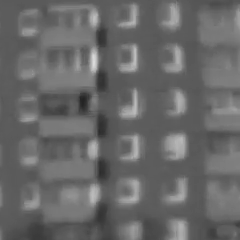}} \hspace{0.01mm}
\subfloat[]{\includegraphics[width=0.32\textwidth]{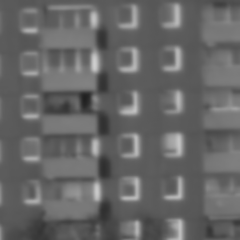}}\hspace{0.01mm} 
\subfloat[]{\includegraphics[width=0.32\textwidth]{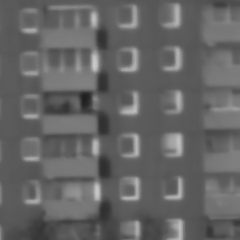}}
\caption{The comparison of reference images obtained by Model 1 and Model 2 from the Building and Car sequences. (a) Observed. (b) Model 1. (c) Model 2. The PSNR of (b), (c) from Car sequence are 28.1232 and 28.2689 (in dB) respectively. The SSIM of (b), (c) from Car sequence are 0.8547 and 0.8624 respectively. The computational times of Model 1 on Building and Car sequence are 1.357 and 1.164 (in seconds) respectively. The computational times of Model 2 on Building and Car sequence are 269.0 and 574.1 (in seconds) respectively. Note that blind deconvolution for deblurring has not been applied to these results.}
\label{fig:model comparsion}
\end{figure}

\subsubsection{Comparison between Model 1 and Model 2}
One aim of our proposed method is to obtain a good restored image. Model 1 gives a fast and reasonable result. The computation time is within 2 seconds in general. Also, the obtained restored image is satisfactory for further usage, for instance, registration purpose. The efficiency of Model 1 owes to the simple 2-norm of the fidelity term. However, if the video is severely distorted or the reference image is required to be of high visual quality, Model 1 may not able to fulfill these aims. It is because the temporal average of a severely distorted video may give a noticeable localized blur on the distorted pixel. Therefore, Model 2 is proposed to tackle this kind of situation, as it has a fidelity term involving the low-rank part of the observed images. This fidelity term can give a more accurate result and mitigate the blurring effect.  

In Figure \ref{fig:model comparsion}, the comparison between Model 1 and Model 2 on both the synthetic and real severely distorted video is illustrated. The PSNR and SSIM of the restored image obtained from Model 2 are slightly higher than that of Model 1. Also, the blurring effect of the restored image in Model 2 is weaker than that in Model 1. For example, the boundary of the windows in the restored image obtained by applying the proposed method with Model 1 in the Building sequence is blurry, while the blurring effect is mitigated in that obtained Model 2. This is because the fidelity term involving low-rank part in the Model 2 gives a more accurate similarity measure. The computational time of Model 2 is much longer than Model 1 as computing RPCA is relatively costly, especially in computing the initial low-rank part of the observed video which usually consists of about 100 frames. Therefore, in general, Model 2 is applied in the severe turbulence-degraded video or more demanding restoration result. 

\begin{figure}[t]
\centering
\subfloat[]{\includegraphics[width=0.24\textwidth]{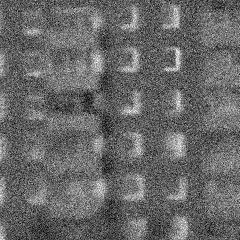}} \hspace{0.01mm}
\subfloat[]{\includegraphics[width=0.24\textwidth]{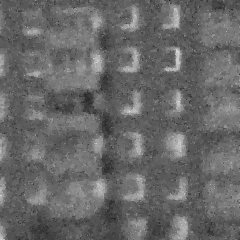}}\hspace{0.01mm} 
\subfloat[]{\includegraphics[width=0.24\textwidth]{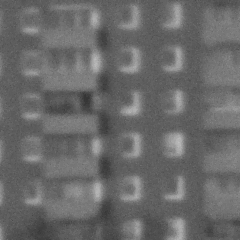}}\hspace{0.01mm}
\subfloat[]{\includegraphics[width=0.24\textwidth]{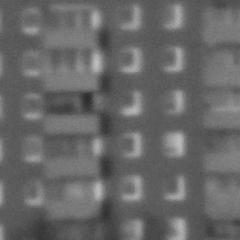}} \\ 
\vspace{-3mm}
\subfloat[]{\includegraphics[width=0.24\textwidth]{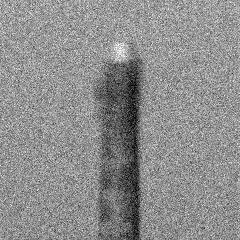}} \hspace{0.01mm}
\subfloat[]{\includegraphics[width=0.24\textwidth]{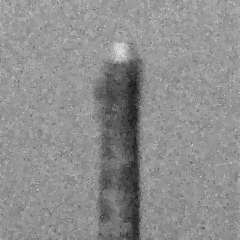}}\hspace{0.01mm} 
\subfloat[]{\includegraphics[width=0.24\textwidth]{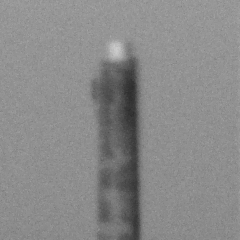}}\hspace{0.01mm}
\subfloat[]{\includegraphics[width=0.24\textwidth]{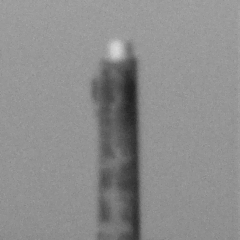}}
\caption{The restored images obtained by Model 3 from the noisy Building and Chimney sequences. (a)(e) Observed. (b)/(f) Denoised (a)/(e). (c)(g) Denoised from the whole sequence. (d)(h) Denoised by Model 3. Note that blind deconvolution for deblurring has not been applied to these results.}
\label{fig:noise}
\end{figure}

\subsubsection{Analysis on Model 3}
Our proposed method has a general setting and gives the flexibility to tackle different problems. For example, the observed turbulence-degraded sequence is severely degraded by noise. Model 3 which consists of a TV regularization term on the restored image is proposed to tackle this problem. To demonstrate the effectiveness of Model 3, Gaussian noise is added to the Building and Chimney sequences, which consists of 30 less noisy and 70 more noisy frames. Also, experiments are carried out with these noisy sequences. In general, it is very hard to have a satisfactory result by denoising one single noisy image if the noise is strong. However, if we have a sequence of noisy images of the same stationary object, a better denoised result can be obtained. Unfortunately, this is not our case as all the images are distorted, and thus the object positions do not align well. As a result, some comparatively good images are needed to subsampled to limit the magnitude of distortion and noise, so as to obtain a satisfactory result.

In Figure \ref{fig:noise}, the restored image obtained by the proposed method with Model 3 is shown. In Figure \ref{fig:noise}(b), the denoising result is not satisfactory as the noise level of the chimney image is strong. If all observed images are taken into account in the denoising model, the restored image is blurry. In Figure \ref{fig:noise}(c), since all observed images are taken into account, including severely distorted images, the resultant image is blurry. If we can subsample those mildly distorted and less noisy images, we can obtain a comparatively sharper result. See Figure \ref{fig:noise}(d).

\section{Conclusion} \label{sec:Conclusion}
This paper presents a general framework to simultaneously restore an image and obtain an optimal subsample consisting of mildly distorted and sharp frames. Also, three models with different fidelity terms and regularization terms are proposed along with the corresponding efficient algorithms. The major tasks are (1) speeding up the restoration of a clear image from turbulence-degraded video, (2) quickly restore a clear image from video severely degraded by turbulence and noise without applying costly image registration techniques. To solve the first task, we propose the IRIS algorithm to alternatively optimize the energy in Model 1, which consists of a simple yet effective $L^2$ fidelity term, and regularizers on image sharpness and subsample size to restore a clear image within 2 seconds for a 100-frames video. To tackle the second task, the LIRIS and TVIRIS algorithms are proposed, which are instead equipped with a low-rank fidelity term and a TV regularization term respectively, to restore an image from a severely turbulence-degraded video with additive Gaussian noise. As a by-product of the proposed algorithm, the restoration of other state-of-the-art methods can also be significantly enhanced by applying the proposed restored image as a reference image and optimal subsampled video as the input observed video in their corresponding algorithms. In the future, we are going to apply the proposed general framework to more applications, such as restoring images from other turbulent medium, and investigating the possibility of other fidelity and regularization terms. 
\bibliographystyle{siamplain}
\bibliography{ref}
\end{document}

%% file: ex_shared.tex
\usepackage{lipsum}
\usepackage{amsfonts}
\usepackage{graphicx}
\usepackage{epstopdf}
\usepackage{algorithmicx}
\usepackage{amsmath,amssymb}
\usepackage{algpseudocode}

  \usepackage{paralist}
  \usepackage{graphics} 
  \usepackage{epsfig} 
\usepackage{graphicx}
\graphicspath{ {images/} }
\usepackage{float}

\usepackage[caption=false]{subfig} 
 \usepackage{epstopdf}
 \usepackage{hyperref}

 \usepackage{multirow}
 \hypersetup{urlcolor=blue, citecolor=red}

\ifpdf
  \DeclareGraphicsExtensions{.eps,.pdf,.png,.jpg}
\else
  \DeclareGraphicsExtensions{.eps}
\fi
\DeclareMathOperator*{\argmin}{\mathbf{argmin}}

\DeclareMathOperator{\shrink}{\mathbf{shrink}}
\numberwithin{theorem}{section}

\newcommand{\TheTitle}{Variational models for joint subsampling and reconstruction of turbulence-degraded images} 
\newcommand{\TheAuthors}{Chun Pong Lau, Yu Hin Lai and Lok Ming Lui}
\newcommand{\ShortTitle}{Joint image reconstruction and image subsampling}  

\headers{\ShortTitle}{\TheAuthors}

\title{{\TheTitle}}
\author{
  Chun Pong Lau\thanks{Department of Mathematics, The Chinese University of Hong Kong, Shatin, Hong Kong (\email{cplau@math.cuhk.edu.hk, yhlai@math.cuhk.edu.hk, lmlui@math.cuhk.edu.hk}).}
  \and
  Yu Hin Lai\footnotemark[1]
  \and
  Lok Ming Lui\footnotemark[1]
}

\usepackage{amsopn}


%% file: main.bbl
\begin{thebibliography}{10}

\bibitem{anantrasirichai2013atmospheric}
{\sc N.~Anantrasirichai, A.~Achim, N.~G. Kingsbury, and D.~R. Bull}, {\em
  Atmospheric turbulence mitigation using complex wavelet-based fusion}, IEEE
  Transactions on Image Processing, 22 (2013), pp.~2398--2408.

\bibitem{aubailly2009automated}
{\sc M.~Aubailly, M.~A. Vorontsov, G.~W. Carhart, and M.~T. Valley}, {\em
  Automated video enhancement from a stream of atmospherically-distorted
  images: the lucky-region fusion approach}, in Proc. SPIE, vol.~7463, 2009,
  p.~74630C.

\bibitem{candes2011robust}
{\sc E.~J. Cand{\`e}s, X.~Li, Y.~Ma, and J.~Wright}, {\em Robust principal
  component analysis?}, Journal of the ACM (JACM), 58 (2011), p.~11.

\bibitem{fried1978probability}
{\sc D.~L. Fried}, {\em Probability of getting a lucky short-exposure image
  through turbulence}, JOSA, 68 (1978), pp.~1651--1658.

\bibitem{furhad2016restoring}
{\sc M.~H. Furhad, M.~Tahtali, and A.~Lambert}, {\em Restoring
  atmospheric-turbulence-degraded images}, Applied optics, 55 (2016),
  pp.~5082--5090.

\bibitem{goldstein2009splitbregman}
{\sc T.~Goldstein and S.~Osher}, {\em The split bregman method for
  l1-regularized problems}, SIAM Journal on Imaging Sciences, 2 (2009),
  pp.~323--343, \url{https://doi.org/10.1137/080725891}.

\bibitem{he2016atmospheric}
{\sc R.~He, Z.~Wang, Y.~Fan, and D.~Fengg}, {\em Atmospheric turbulence
  mitigation based on turbulence extraction}, in Acoustics, Speech and Signal
  Processing (ICASSP), 2016 IEEE International Conference on, IEEE, 2016,
  pp.~1442--1446.

\bibitem{hirsch2010efficient}
{\sc M.~Hirsch, S.~Sra, B.~Sch{\"o}lkopf, and S.~Harmeling}, {\em Efficient
  filter flow for space-variant multiframe blind deconvolution}, in Computer
  Vision and Pattern Recognition (CVPR), 2010 IEEE Conference on, IEEE, 2010,
  pp.~607--614.

\bibitem{hufnagel1964modulation}
{\sc R.~Hufnagel and N.~Stanley}, {\em Modulation transfer function associated
  with image transmission through turbulent media}, JOSA, 54 (1964),
  pp.~52--61.

\bibitem{li2007atmospheric}
{\sc D.~Li, R.~M. Mersereau, and S.~Simske}, {\em Atmospheric
  turbulence-degraded image restoration using principal components analysis},
  IEEE Geoscience and Remote Sensing Letters, 4 (2007), pp.~340--344.

\bibitem{lin2010augmented}
{\sc Z.~Lin, M.~Chen, and Y.~Ma}, {\em The augmented lagrange multiplier method
  for exact recovery of corrupted low-rank matrices}, arXiv preprint
  arXiv:1009.5055,  (2010).

\bibitem{lou2013video}
{\sc Y.~Lou, S.~H. Kang, S.~Soatto, and A.~L. Bertozzi}, {\em Video
  stabilization of atmospheric turbulence distortion.}, Inverse Problems \&
  Imaging, 7 (2013).

\bibitem{meinhardt2014implementation}
{\sc E.~Meinhardt-Llopis and M.~Micheli}, {\em Implementation of the centroid
  method for the correction of turbulence}, Image Processing On Line, 4 (2014),
  pp.~187--195.

\bibitem{micheli2014linear}
{\sc M.~Micheli, Y.~Lou, S.~Soatto, and A.~L. Bertozzi}, {\em A linear systems
  approach to imaging through turbulence}, Journal of mathematical imaging and
  vision, 48 (2014), pp.~185--201.

\bibitem{pearson1976atmospheric}
{\sc J.~E. Pearson}, {\em Atmospheric turbulence compensation using coherent
  optical adaptive techniques}, Applied optics, 15 (1976), pp.~622--631.

\bibitem{roggemann1994image}
{\sc M.~C. Roggemann, C.~A. Stoudt, and B.~M. Welsh}, {\em Image-spectrum
  signal-to-noise-ratio improvements by statistical frame selection for
  adaptive-optics imaging through atmospheric turbulence}, Optical Engineering,
  33 (1994), pp.~3254--3265.

\bibitem{roggemann1996imaging}
{\sc M.~C. Roggemann, B.~M. Welsh, and B.~R. Hunt}, {\em Imaging through
  turbulence}, CRC press, 1996.

\bibitem{rudin1992nonlinear}
{\sc L.~I. Rudin, S.~Osher, and E.~Fatemi}, {\em Nonlinear total variation
  based noise removal algorithms}, Physica D: Nonlinear Phenomena, 60 (1992),
  pp.~259--268.

\bibitem{seitz2009filter}
{\sc S.~M. Seitz and S.~Baker}, {\em Filter flow}, in Computer Vision, 2009
  IEEE 12th International Conference on, IEEE, 2009, pp.~143--150.

\bibitem{shimizu2008super}
{\sc M.~Shimizu, S.~Yoshimura, M.~Tanaka, and M.~Okutomi}, {\em
  Super-resolution from image sequence under influence of hot-air optical
  turbulence}, in Computer Vision and Pattern Recognition, 2008. CVPR 2008.
  IEEE Conference on, IEEE, 2008, pp.~1--8.

\bibitem{lrslibrary2015}
{\sc A.~Sobral, T.~Bouwmans, and E.-h. Zahzah}, {\em Lrslibrary: Low-rank and
  sparse tools for background modeling and subtraction in videos}, in Robust
  Low-Rank and Sparse Matrix Decomposition: Applications in Image and Video
  Processing, CRC Press, Taylor and Francis Group., 2015.

\bibitem{tyson2015principles}
{\sc R.~K. Tyson}, {\em Principles of adaptive optics}, CRC press, 2015.

\bibitem{vorontsov1999parallel}
{\sc M.~A. Vorontsov}, {\em Parallel image processing based on an evolution
  equation with anisotropic gain: integrated optoelectronic architectures},
  JOSA A, 16 (1999), pp.~1623--1637.

\bibitem{vorontsov2001anisoplanatic}
{\sc M.~A. Vorontsov and G.~W. Carhart}, {\em Anisoplanatic imaging through
  turbulent media: image recovery by local information fusion from a set of
  short-exposure images}, JOSA A, 18 (2001), pp.~1312--1324.

\bibitem{xie2016removing}
{\sc Y.~Xie, W.~Zhang, D.~Tao, W.~Hu, Y.~Qu, and H.~Wang}, {\em Removing
  turbulence effect via hybrid total variation and deformation-guided kernel
  regression}, IEEE Transactions on Image Processing, 25 (2016),
  pp.~4943--4958.

\bibitem{zhu2013removing}
{\sc X.~Zhu and P.~Milanfar}, {\em Removing atmospheric turbulence via
  space-invariant deconvolution}, IEEE transactions on pattern analysis and
  machine intelligence, 35 (2013), pp.~157--170.

\end{thebibliography}
